\documentclass[a4paper,10pt]{article}
\pdfoutput=1
\usepackage{ifthen}
\newboolean{ieee}
\setboolean{ieee}{false}
\newboolean{review}
\setboolean{review}{false}

\usepackage{afterpage}

\usepackage[pdfpagemode=none,pdfstartview=FitH,pdfhighlight=/O]{hyperref}
\hypersetup{colorlinks=true,citecolor=blue,urlcolor=blue,linkcolor=blue,menucolor=blue}
\hypersetup{bookmarks=true,bookmarksopen=true}

\usepackage{hypernat} 
\usepackage{datetime} 
\usepackage{graphicx} 
\usepackage{epstopdf} 
\usepackage{subfigure} 
\usepackage[bf]{caption} 
\usepackage{ifpdf} 
\usepackage{color}
\usepackage[pstarrows]{pict2e} 
\usepackage{lineno}
\usepackage{multirow}
\ifreview
\fi
\ifieee
\usepackage[square,comma,numbers,sort]{natbib}
\else
\usepackage[round,colon,authoryear,sort&compress]{natbib}
\usepackage[a4paper,portrait]{geometry}
\makeatletter
\renewcommand{\@thesubfigure}{{\subcaplabelfont(\alph{subfigure})}\space}
\makeatother

\fi

\usepackage{rotating}
\usepackage{times}
\usepackage{xspace}

\newcommand{\etc}{{\it etc.}\xspace}
\newcommand{\ie}{{\it i.e.},\xspace}
\newcommand{\eg}{{\it e.g.},\xspace}

\newcommand{\dson}{\renewcommand{\baselinestretch}{1.5}\large\normalsize}
\newcommand{\dsoff}{\renewcommand{\baselinestretch}{1}\large\normalsize}

\let\orgautoref\autoref
\providecommand{\Autoref}
        {\def\equationautorefname{Equation}%
         \def\figureautorefname{Figure}%
         \def\subfigureautorefname{Figure}%
         \def\sectionautorefname{Section}%
         \def\subsectionautorefname{Section}%
         \def\subsubsectionautorefname{Section}%
         \def\Itemautorefname{Item}%
         \def\tableautorefname{Table}%
         \orgautoref}

\renewcommand{\autoref}
        {\def\equationautorefname{eq.}%
         \def\figureautorefname{fig.}%
         \def\subfigureautorefname{fig.}%
         \def\sectionautorefname{sect.}%
         \def\subsectionautorefname{sect.}%
         \def\subsubsectionautorefname{sect.}%
         \def\Itemautorefname{item}%
         \def\tableautorefname{tab.}%
         \orgautoref}
\providecommand{\autorefs}
        {\def\equationautorefname{eqs.}%
         \def\figureautorefname{figs.}%
         \def\subfigureautorefname{figs.}%
         \def\sectionautorefname{sects.}%
         \def\subsectionautorefname{sects.}%
         \def\subsubsectionautorefname{sects.}%
         \def\Itemautorefname{items}%
         \def\tableautorefname{tabs.}%
         \orgautoref}

\newcommand{\X}{\mathbf{X}}
\newcommand{\E}{\mathbf{E}}
\newcommand{\Y}{\mathbf{Y}}
\newcommand{\R}{\mathbf{R}}
\newcommand{\w}{\mathbf{w}}
\renewcommand{\v}{\mathbf{v}}

\newcommand{\pcbc}{DIM\xspace} 
\newcommand{\alg}[1]{\texttt{#1}}
\newcommand{\tabind}{\hspace*{3mm}}
  
\begin{document}



\ifieee
\title{Explaining Away Results in Accurate and Tolerant Template Matching}
\author{Michael W. Spratling}
\IEEEcompsocitemizethanks{\IEEEcompsocthanksitem Michael Spratling is with the Department of Informatics, King's College London, London. UK.}
\else
\noindent {\Large \bf Explaining Away Results in Accurate and Tolerant Template Matching}\\\par
\noindent {\large \bf M. W. Spratling}\\
\noindent {\normalsize King's College London, Department of Informatics, London. UK. michael.spratling@kcl.ac.uk}\\

\ifreview
\noindent \begin{tabular}{@{}rl}
Correspondence should be addressed to: &M. W. Spratling\\
&Department of Informatics\\&King's College London\\
&Strand Campus\\
&Bush House\\
&30 Aldwych\\&London WC2B 4BG\\&UK\\
email: &michael.spratling@kcl.ac.uk\\telephone:&+44 20 7848 2027
\end{tabular}

\paragraph{ORCID ID:} 0000-0001-9531-2813
\paragraph{Submitted:} \today

\newpage
\dson
\fi
\fi
\ifieee\IEEEtitleabstractindextext{%
\fi
  \begin{abstract}  
\noindent
  Recognising and locating image patches or sets of image features is an
  important task underlying much work in computer vision. Traditionally this has
  been accomplished using template matching. However, template matching is
  notoriously brittle in the face of changes in appearance caused by, for
  example, variations in viewpoint, partial occlusion, and non-rigid
  deformations. This article tests a method of template matching that is more
  tolerant to such changes in appearance and that can, therefore, more
  accurately identify image patches.  In traditional template matching the
  comparison between a template and the image is independent of the other
  templates. In contrast, the method advocated here takes into account the
  evidence provided by the image for the template at each location and the full
  range of alternative explanations represented by the same template at other
  locations and by other templates. Specifically, the proposed method of
  template matching is performed using a form of probabilistic inference known
  as ``explaining away''.  The algorithm used to implement explaining away has
  previously been used to simulate several neurobiological mechanisms, and been
  applied to image contour detection and pattern recognition tasks.  Here is is
  applied for the first time to image patch matching, and is shown to produce
  superior results in comparison to the current state-of-the-art methods.
\end{abstract}
\ifieee
}
\maketitle
\setcounter{page}{1}
\else
\ifreview\newpage\fi
\fi


\section{Introduction}
\label{sec-introduction}

Recognising that one part of an image contains a particular object, image
structure or set of local image features is a fundamental sub-problem in many
image processing and computer vision algorithms. For example, it can be used to
perform object detection or recognition by identifying parts belonging to an
object category
\citep{Lowe99,Leibe_etal08,Csurka_etal04}, or for
navigation by identifying and locating landmarks in a scene
\citep{Ozuysal_etal10,Se_etal05}, or for image mosaicing/stitching by
identifying corresponding locations in multiple images
\citep{Szeliski06,BrownLowe07}. Similarly, extracting a distinctive image
structure in one image and then recognising and locating that same feature in
another image of the same scene taken from a different viewpoint, or at a
different time, is essential to solving the stereo and video correspondence
problems, and hence, for calculating depth or motion, and for performing image
registration and tracking
\citep{LucasKanade81,Brown_etal03,Gall_etal11,ZitovaFlusser03}. Finally,
locating specific image features is fundamental to tasks such as edge detection
\citep{Canny86}. In this latter case the image structure being searched for is
usually defined mathematically (\eg a Gaussian derivative for locating intensity
discontinuities), whereas for solving image stitching and correspondence problems the image
structure being searched for is extracted from another image, and in the case of
object or landmark recognition the target image features may have been learnt
from a set of training images.

Image patch recognition has traditionally been accomplished using template
matching \citep{Ma_etal09,WeiLai08,Goshtasby_etal84,BarneaSilverman72}. In this
case the image structure being searched for is represented as an array of pixel
intensity values, and these values are compared with the pixel intensities
around every location in the query image. There are many different metrics that
can be used to quantify how well the template matches each location in an image;
such as the sum of absolute differences (SAD), the normalised cross-correlation
(NCC), the sum of squared differences (SSD) or the zero-mean normalised cross
correlation\footnote{Also known as the sample Pearson correlation coefficient.}
(ZNCC). For any metric it is necessary to define a criteria that needs to be met
for a patch of image to be considered a match to the template. For
correspondence problems, it is often assumed that the image structure being
searched-for will match exactly one location in the query image. The location
with the highest similarity to the template is thus considered to be the
matching location. However, additional criteria, such as the ratio of the two
highest similarity values, may be used to reject some such putative matches. In other
tasks, the image structure being searched for may occur zero, one or multiple
times in the query image. In this case it is necessary to define a global
threshold to distinguish those image locations where the template matches the
query image from those locations where it does not. It is also typically the
case that for a patch of image to be considered a match to the template it must
be more similar to the template than its immediate neighbours. Hence, locations
where the template is considered to match the image are ones where the
similarity metric is a local maximum and exceeds a global threshold.

With traditional template matching, because the metric for assessing the
similarity of the template and a patch of image compares intensity values, the
result can be effected by changes in illumination. This issue can be resolved by
using a metric such as ZNCC which subtracts the mean intensity from the template
and from the image patch it is being compared to. This results in a comparison
of the relative intensity values and produces a similarity measure that is
tolerant to differences in lighting conditions between the query image and the template.

Another issue is that the metric for assessing the similarity of a template to a
patch of image is based on pixel-wise comparisons of intensity values. The
results will therefore be effected if the pixels being compared do not
correspond to the same part of the target image structure. This problem can
arise due to differences in the appearance of the image structure between the
query image and the template, caused by variations in viewpoint, partial
occlusion, and non-rigid deformations. To be able to recognise image features
despite such changes in appearance one approach is to use multiple templates
that represent the searched-for image patch across the range of expected changes
in appearance. However, even small differences in scale, orientation, aspect
ratio, \etc can result in sufficient mis-alignment at the pixel-level that a low
value of similarity is calculated by the similarity metric for all
transformations of the template when compared to the correct location in the
query image. Hence, to allow tolerance to viewpoint, even when using multiple
templates for each image feature, it is necessary to use a low threshold to
avoid excluding true matches. However, given the large number of comparisons
that are being made between all templates and all image locations, a low
threshold will inevitably lead to false-positives.  There is thus an
irreconcilable need both for a high threshold to avoid false matches and for a
low threshold in order not to exclude true matches in situations where the
template is not perfectly aligned with the image.  These problems have lead to
template matching being abandoned in favour of alternative methods (see
\autoref{sec-review}) for most tasks except for low-level ones such as edge
detection.

This article shows that the performance of template matching can be
significantly improved by requiring templates to compete with each other to
match the image. The particular type of competition used in the proposed method,
called Divisive Input Modulation \citep[DIM][]{Spratling_etal09,Spratling17a},
implements a form of probabilistic inference known as ``explaining away''
\citep{Kersten_etal04,LochmannDeneve11}. This
means that the similarity between a template and a patch of image takes into
account not only the similarity in the pixel intensity values at corresponding
locations in the template and the patch, but also the range of alternative
explanations for the patch intensity values represented by the same template at
other locations and by other templates. If the similarity between a template and
each image location is represented by an array, then this array is dense for
traditional template matching. In contrast, due to the competition employed by
the proposed method, the array of similarity values is very sparse. Those
locations that match a template can therefore be more readily identified and
there is typically a much larger range of threshold values that separate true
matches from false matches.

\section{Related Work}
\label{sec-review}

Given the issues with template matching discussed above, many alternative
methods for locating image features have been developed. Typically, these
alternative methods change the way the template and image patch are represented,
so that the comparison is performed in a different feature-space, or change the
computation that is used to perform the comparison, or use a combination of
both.

One alternative is to employ a classifier in place of the comparison of
corresponding pixel intensity values used in traditional template matching. For
example, random trees and random ferns can be trained using image patches seen
from multiple viewpoints in order to robustly recognise those image features
when they appear around keypoints extracted from a new image
\citep{Gall_etal11,Ozuysal_etal10,LepetitFua06}. Sliding-window
based methods apply the classifier, sequentially, to all regions within the
image \citep{DalalTriggs05,Lampert_etal08}, while region-based methods select a
sub-set of image patches for presentation to the classifier
\citep{Girshick_etal16,Gu_etal09,Uijlings_etal13}.  In each case, the classifier
provides robustness to changes in appearance due to, for example, viewpoint or
within-class variation. Further tolerance to appearance can be achieved by using
windows with different sizes and aspect ratios.  A classifier in the form of a
deep neural network can also be used to directly assess the similarity between
two image patches \citep{ZagoruykoKomodakis15,ZagoruykoKomodakis17}.

Instead of being used to perform the comparison between a template and an image
patch, a deep neural network can also be used to extract features from the image
and template for comparison (\ie a deep neural network can be used to change the
feature-space, rather than change the similarity computation). For example,
\citet{Kim_etal17} used a convolutional neural network (CNN) to represent both
the template and the image in a new feature-space, the comparison was then
carried out using NCC.

Histogram matching is another method that changes the feature-space. Histogram
matching compares the colour histograms of the template and image patch, and
hence, disregards all spatial information \citep{Comaniciu_etal00}. This will
introduce tolerance to differences in appearance, but also reduces the ability
to discriminate between spatially distinct features.  Co-occurrence based
template matching (CoTM) calculates the cost of matching a template to an image
patch as inversely proportional to the probability of the corresponding pixel
values co-occurring in the image \citep{Kat_etal18}. This can be achieved by
mapping the points in the image and template to a new feature-spaced defined by
the co-occurrence statistics. However, this method does not work well for
grayscale images or images containing repeating texture, and is not tolerant to
differences in illumination \citep{Kat_etal18}.

Another approach to is to perform comparisons on more distinctive image features
than image intensity values. For example, the scale-invariant feature transform
(SIFT) generates an image descriptor that is invariant to illumination,
orientation, and scale and partially invariant to affine distortion
\citep{Lowe99,Lowe04}. Methods to allow SIFT descriptors to be matched across
images with invariance to affine transformations have also been developed
\citep{MorelYu09,DongSoatto15}.
Many alternative feature descriptors have also been proposed,
such as SURF \citep{Bay_etal06}, BRIEF \citep{Calonder_etal10},
ORB \citep{Rublee_etal11}, GLOH \citep{MikolajczykSchmid05}, DAISY
\citep{Tola_etal08}, and BINK \citep{Saleiro_etal17}. However, experiments
comparing the performance of different image descriptors for finding matching
locations between images of the same scene suggest that SIFT remains one of the
most accurate methods
\citep{Balntas_etal18,MikolajczykSchmid05,Wu_etal13,TareenSaleem18,Balntas_etal17b,Mukherjee_etal15,Khan_etal15}.
It is also possible to learn image descriptors, and this approach can improve
performance beyond that of hand-crafted descriptors
\citep{Trzcinski_etal12,Brown_etal11,Simonyan_etal14,Schonberger_etal17}.
Recently, learning image descriptors using deep neural networks has become a
popular approach
\citep{Kwang_etal16,Simo-Serra_etal15,Balntas_etal18,Balntas_etal17,Balntas_etal16,ZagoruykoKomodakis15,ZagoruykoKomodakis17,Mitra_etal17}.

Another alternative to traditional template matching, that can perform image
patch matching with tolerance to changes in appearance, is image alignment. In
these methods the aim is to find the affine transformation that will align a
template with the image \citep{LucasKanade81,ZhangAkashi15,Korman_etal13}. For
example, FAsT-Match is a relatively recent algorithm of this type that measures
the similarity between a template and an image patch by first searching for the
2D affine transformation that maximises the pixel-wise similarity
\citep{Korman_etal13}. However, it is limited to working with grayscale images
and the large search-space of possible affine transformations makes this
algorithm slow. A more recent variation on this algorithm, OATM
\citep{Korman_etal18}, has increased speed but remains both slower and less
accurate than another approach, DDIS, which is discussed in the following
paragraph.

Another approach is to define alternative metrics for comparing the template
with the image that allow for mis-alignment between the pixels in the template
and the corresponding pixels in the image, rather than rigidly comparing pixels
at corresponding locations in the template and the image. Typically, these
metrics are based on measuring the distance between points in the template and
the best matching points in the image
\citep{Talmi_etal17,Dekel_etal15,Oron_etal18}. 
For example, the Best-Buddies Similarity (BBS) metric
\citep{Dekel_etal15,Oron_etal18} is computed by counting the proportion of
sub-regions in the template and the image patch that are ``best-buddies''. For each
sub-region in the template the most similar sub-region (in terms of position and
colour) is found in the image patch. For each sub-region in the image patch the
most similar sub-region in the template is calculated in the same way. A pair of
sub-regions are best-buddies if they are most similar to each other. Sub-regions
can be best-buddies even if they are not at corresponding locations in the
template and the image patch, and this thus provides tolerance to differences in
appearance between the template and the patch.  Deformable diversity similarity
(DDIS) is similar to BBS, but it differs in the way it deals with spatial
deformations, and the criteria used for determining if sub-regions in the image
patch and template match \citep{Talmi_etal17}. Specifically, for every
sub-region of the image patch the most similar sub-region (in terms of colour
only) is found in the template. The contribution of each such match to the
overall similarity is inversely weighted by the number of other sub-regions that
have been matched to the same location in the template, and by the spatial
distance between the matched sub-regions in the image patch and the
template. DDIS produces the current state-of-the-art performance when applied to
template matching in colour-feature space on standard benchmarks
\citep{Talmi_etal17,Kat_etal18}.

This article proposes another alternative method of image patch matching. Like
traditional template matching, the proposed method compares pixel-intensity
values in the template with those at corresponding locations in the image patch.
However, in contrast to traditional template matching the similarity between any
one template and the image patch is not independent of the other templates.
Instead, the templates compete with each other to be matched to the image
patch. This article describes empirical tests of this new approach to template
matching that demonstrate that it provides tolerance to differences in
appearance between the template and the same image feature in the query
image. This results in more accurate identification of features in an image
compared to both traditional template matching and recent state-of-the-art
alternatives to template matching \citep{Talmi_etal17,Dekel_etal15,Oron_etal18,Kat_etal18,Kim_etal17,ZagoruykoKomodakis15,ZagoruykoKomodakis17}.

\section{Methods}
\label{sec-methods}

\subsection{Image Pre-processing and Template Definition}
\label{sec-methods_preproc}

Image features are better distinguished using relative intensity (or contrast)
rather than absolute intensity. For this reason, ZNCC is a sensible choice of
similarity metric for template matching. ZNCC subtracts the mean intensity from
the template and from the image patch it is being compared to.  Subtracting the
mean intensity will obviously result in positive and negative relative intensity
values. However, non-negative inputs are required by the mechanism that is used
in this article to implement template matching using explaining
away\footnote{This method is derived \citep{Spratling_etal09} from the version
  of non-negative matrix factorisation \citep[NMF][]{LeeSeung01,LeeSeung99} that
  minimises the Kullback-Leibler (KL) divergence between the input and a
  reconstruction of the input created by the additive combination of elementary
  image components (see \autoref{sec-methods_matching}). Because it minimises
  the KL divergence it requires the input to be non-negative.  Reconstructing
  image data through the addition of image components, as occurs in NMF and in
  the proposed algorithm, is considered an advantage as it is consistent with
  the image formation process in which image components are added together (and
  not subtracted) in order to generate images
  \citep{Beyeler_etal19,LeeSeung01,LeeSeung99,Hoyer04}.  In previous work the
  algorithm used here to implement template matching using explaining away has
  been used to simulate the response properties of neurons in the primary visual
  cortex \citep{Spratling11a}.  In biological neural networks, variables are
  represented by firing rates which can not be negative. Hence, in these
  previous applications the restriction to working with non-negative values had
  the advantage of increasing the biological-plausibility of the model. In this
  context, the pre-processing defined in this section can be considered to be a
  simple model of the processing that is performed in the retina to generate ON
  and OFF responses that respectively signal increases and decreases in
  brightness.  } (see \autoref{sec-methods_matching}). Hence, to allow the
proposed method to process relative intensity values the input image was
pre-processed as follows.

A grayscale input image $I$ was convolved with a 2D circular-symmetric Gaussian
mask $g$ with standard deviation equal to $\sigma$ pixels, such that: $\bar{I}=I
\ast g$.  $\bar{I}$ is an estimate of the local mean intensity across the
image. To avoid a poor estimate of $\bar{I}$ near the borders, the image was
first padded on all sides with intensity values that were mirror reflections of
the image pixel values near the borders of $I$. The width of the padding was
equal to the width of the template on the left and right borders, and equal to
the height of the template on the the top and bottom borders. Once calculated
$\bar{I}$ could be cropped to be the same size as the original input
image. However, to avoid edge-effects when template matching, $\bar{I}$ was left
padded and all the arrays the same size as $\bar{I}$ (\ie $\X$, $\R$, $\E$, and
$\Y$ see \autoref{sec-methods_matching}) were cropped to be the same size as the
original image once the template matching method had been applied\footnote{An
  alternative approach to avoid edge effects is to set to zero the similarity
  values near to the borders of the image. This alternative approach is employed
  in other patch matching algorithms such as BBS and DDIS, as can be seen in the
  4th and 5th rows of \autoref{fig-bss_data_examples}. This alternative approach
  has the advantage of increasing the processing speed, as similarity values do
  not need to be calculated for regions of the image adjacent to the edges, but
  has the disadvantage that it may prevent detection of the image patch
  corresponding to the target if it is very close to the border of the
  image.}. The relative intensity can be approximated as $\X=2(I-\bar{I})$. To
produce only non-negative input to the proposed method, the positive and
rectified negative values of $\X$ were separated into two images $\X_1$ and
$\X_2$. Hence, for grayscale images the input to the model was two arrays
representing increases and decreases in local contrast. For colour images each
colour channel was pre-processed in the same way, resulting in six input arrays
($\X_1 \dots \X_6$) representing the increases and decreases about the average
value in each colour channel.

For grayscale images a template consists of two arrays of values ($\w_{j1}$
and $\w_{j2}$) which are compared to $\X_1$ and $\X_2$. Similarly for a
colour image a template consists of six arrays of values ($\w_{j1} \dots
\w_{j6}$).  These arrays can be produced by performing the pre-processing
operation described in the previous paragraph on a standard template of
intensity values (which could have been defined mathematically, have been
learnt, or have been a patch extracted from an image). Alternatively, the
templates can be created by extracting regions from images that have been
processed as described in the preceding paragraph. This latter method was used
in all the experiments reported in this article. The value of $\sigma$ was set
equal to half of the template width or height, whichever was the smaller of the
two dimensions.

\subsection{Template Matching}
\label{sec-methods_matching}

The proposed method of template matching was implemented using the Divisive Input
Modulation (DIM) algorithm \citep{Spratling_etal09}. This algorithm has been
used previously to simulate neurophysiological \citep[][]{Spratling11a} and
psychological data \citep[][]{Spratling16b} and applied to tasks in robotics
\citep[][]{MuhammadSpratling15}, pattern recognition \citep[][]{Spratling14b}
and computer vision \citep[][]{Spratling13a,Spratling17a}. 
The DIM algorithm is described in
these previous publications, but this description is repeated here for the
convenience of the reader. DIM was implemented using the following equations:
\begin{equation}
  \R_i= \sum_{j=1}^{p} \left(\v_{ji} \ast \Y_j\right)
\label{eq-pcbc_r}
\end{equation}
\begin{equation}
  \E_i=\X_i \oslash \left[\R_i\right]_{\epsilon_2}
  \label{eq-pcbc_e}
\end{equation}
\begin{equation}
  \Y_j \leftarrow \left[\Y_j\right]_{\epsilon_1} \odot \sum_{i=1}^{k} \left(\w_{ji} \star \E_i\right)
\label{eq-pcbc_y}
\end{equation}
Where $i$ is the index over the number of input channels (the maximum index $k$
is two for grayscale images and six for colour images), $j$ is an index over the
number, $p$, of different templates being compared to the image; $\X_i$ is a
2-dimensional array generated from the original image by the pre-processing
method described in \autoref{sec-methods_preproc}; $\R_i$ is a 2-dimensional
array representing a reconstruction of $\X_i$; $\E_i$ is a 2-dimensional array
representing the discrepancy (or residual error) between $\X_i$ and $\R_i$; $\Y_j$ is a 2-dimensional
array that represent the similarity between template $j$ and the image at each
pixel; $\w_{ji}$ is a 2-dimensional array representing channel $i$ of template
$j$ defined as described in \autoref{sec-methods_preproc}; $\v_{ji}$ is a
2-dimensional array also representing template values ($\v_{ji}$ and $\w_{ji}$
differ only in the way they are normalised, as described below); $\left[
  v\right]_{\epsilon}=\alg{max}(\epsilon,v)$; $\epsilon_1$ and $\epsilon_2$ are
parameters; $\oslash$ and $\odot$ indicate element-wise division and
multiplication respectively; $\star$ represents cross-correlation; and $\ast$
represents convolution (which is equivalent to cross-correlation with the kernel
rotated $180^o$).

DIM attempts to find a sparse set of elementary components that when combined
together reconstruct the input with minimum error \citep{Spratling14b}. For the
current application, the elementary components are the templates reproduced at every
location in the image, and all templates at all locations can be thought of as a
``dictionary'' or ``codebook'' that can be used to reconstruct many different
inputs.  The activation dynamics, described by \autorefs{eq-pcbc_r},
\ref{eq-pcbc_e} and~\ref{eq-pcbc_y}, perform gradient descent on the residual
error in order to find values of $\Y$ that accurately reconstruct the current
input \citep{Achler13,Spratling_etal09,Spratling_dim-learning}. Specifically,
the equations operate to find values for $\Y$ that minimise the Kullback-Leibler
divergence between the input ($\X$) and the reconstruction of the input ($\R$)
\citep{Spratling_etal09,SolbakkenJunge11}.  The activation dynamics thus result
in the \pcbc algorithm selecting a subset of dictionary elements that best
explain the input. The strength of an element in $\Y$ reflects the strength with
which the corresponding dictionary entry (\ie template) is required to be
present in order to accurately reconstruct the input at that location.

Each element in the similarity array $\Y$ can be considered to represents a
hypothesis about the image features present in the image, and the input $\X$
represents sensory evidence for these different hypotheses. Each similarity
value is proportional to the belief in the hypothesis represented, \ie the
belief that the image features represented by that template are present at that
location in the image.  If a template and a patch of image have strong
similarity this will inhibit the inputs being used to calculate the similarity
of the same template at nearby locations (ones where the templates overlap
spatially), and will also inhibit the inputs being used to calculate the
similarity of other templates at the same and nearby locations.  Thus,
hypotheses that are best supported by the sensory evidence inhibit other
competing hypotheses from receiving input from the same evidence. Informally we
can imagine that overlapping templates inhibit each other's inputs. This
generates a form of competition between templates, such that each one
effectively tries to block other templates from responding to the pixel
intensity values which it represents
\citep{Spratling_etal09,Spratling_dim-learning}. This competition between the
templates performs explaining away
\citep{Kersten_etal04,LochmannDeneve11,Spratling_dim-learning,Spratling_etal09}. If
a template wins the competition to respond to (\ie have a high similarity to) a
particular pattern of inputs, then it inhibits other templates from responding
to those same inputs. Hence, if one template explains part of the evidence (\ie
a patch of image), then support from this evidence for alternative hypotheses
(\ie templates) is reduced, or explained away.

The sum of the values in each template, $\w_j$, was normalised to sum to
one. The values of $\v_j$ were made equal to the corresponding values of $\w_j$,
except they were normalised to have a maximum value of one.
The cross-correlation operator used in equation~\ref{eq-pcbc_y} calculates the
similarity, $\Y$, for the same set of templates, $\w$, at every pixel location in
the image.  The convolution operation used in equation~\ref{eq-pcbc_r}
calculates the reconstructions, $\R$, for the same set of templates, $\v$, at every
pixel location in the image. The rotation of the kernel performed by convolution
ensures that each channel of the reconstruction $\R_i$ can be compared
pixel-wise to the actual input $\X_i$.

For all the experiments described in this paper (except those exploring
parameter sensitivity reported in \autoref{sec-parameter_sensitivity})
$\epsilon_1$ and $\epsilon_2$ were given the values
$\frac{\epsilon_2}{max\left(\sum_j v_{ji}\right)}$ and $1\times 10^{-2}$
respectively. Parameter $\epsilon_1$ allows elements of $\Y$ that are equal to
zero, to subsequently become non-zero.
Parameter $\epsilon_2$ prevents division-by-zero errors and determines the
minimum strength that an input is required to have in order to effect the values
of $\Y$.  As in all previous work with DIM, these parameters have been given
small values compared to typical values of $\Y$ and $\R$, such that they have
negligible effects on the steady-state values of $\R$, $\E$ and $\Y$. To
determine these steady-state values, all elements of $\Y$ were initially set to
zero, and \autorefs{eq-pcbc_r} to~\ref{eq-pcbc_y} were then iteratively updated
with the new values of $\Y$ calculated by \autoref{eq-pcbc_y} substituted into
\autorefs{eq-pcbc_r} and~\ref{eq-pcbc_y}. This iterative process was terminated
after 10 iterations for the experiments reported in
\autorefs{sec-correspondence_bbs} and \ref{sec-correspondence_vgg} (where the
number of templates varied between 1 and 31) and 20 iterations for the
experiments reported in \autoref{sec-template_matching_vgg} (where 70 templates
were used). It is necessary to increase the number of iterations used as the
number of templates increases as the competition between the templates takes
longer to be resolved. However, for a fixed number of templates the results were
not particularly sensitive to the exact number of iterations used (see
\autoref{sec-parameter_sensitivity}).  The values in array $\Y_j$ produced at
the end of the iterative process were used as a measure of the similarity
between template $j$ and the input image over all spatial locations.

\subsection{Post-Processing}
\label{sec-methods_postproc}

While the similarity array $\Y_j$ for template $j$ is sparse, it is not always
the case that the best matching location is represented by a single element with
a large value. Often the best matching location will be represented by a small
population of neighbouring elements with high values. The size of this
population is usually proportional to the size of the template. To sum the
similarity values within neighbourhoods the similarity array produced by each
template was convolved with a binary-valued kernel that contained ones within an
elliptically shaped region with width and height equal to $\lambda$ times the
width and height of the template. The size of the region over which values were
summed was restricted to be at least one pixel, so that a small $\lambda$ and/or
a small template size did not result in a summation of zero pixels, but would
instead result in an output that was the same as the original $\Y_j$. A value of
$\lambda=0.025$ was used in all experiments unless otherwise stated.  However,
the results were not particularly sensitive to the value of $\lambda$ and
similar results were obtained with a range of different values (see
\autoref{sec-parameter_sensitivity}).

\subsection{Implementation}

Open-source software, written in MATLAB, which performs all the experiments
described in this article is available for download from:
\url{http://www.corinet.org/mike/Code/dim_patchmatching.zip}. This code
compares the performance of DIM to that of several other methods.  Firstly,
the BBS method \citep{Dekel_etal15,Oron_etal18} which was implemented using the
code supplied by the authors of BBS\footnote{\url{http://people.csail.mit.edu/talidekel/Code/BBS_code_and_data_release_v1.0.zip}}. Secondly,
the DDIS algorithm \citep{Talmi_etal17} which was implemented using the code
provided by the authors of DDIS\footnote{\url{https://github.com/roimehrez/DDIS}}.
Finally, traditional template matching using ZNCC as the similarity metric which
was implemented using the MATLAB command normxcorr2. This command is part of the
MATLAB Image Processing Toolbox. 

As described in \autoref{sec-methods_preproc}, the proposed method can be
applied to grayscale or colour images. For colour images, the best results were
found using images in CIELab colour space.  For a fair comparison, all other
algorithms were also applied to colour images. Specifically, BBS was
applied to RGB images \citep[as in][]{Dekel_etal15}, DDIS was also applied to RGB
images \citep[as in][]{Talmi_etal17}. ZNCC was applied to HSV images as this was
found to produce better results than either RGB or CIELab.
To apply ZNCC to colour images the similarity
values were calculated using ZNCC independently for each colour channel, and
then these values were summed to produce the final measure of similarity.

\section{Results}
\label{sec-results}

\subsection{Correspondence using the Best Buddies Similarity Benchmark}
\label{sec-correspondence_bbs}

This section describes an experiment in which a template is extracted from one
image and is used to find the best matching location in a second image of the
same scene. Specifically, each image pair consists of two frames from a
video. Within each image the bounding-box of a target object has been defined
manually. Between the images in each pair, the image patch corresponding to the
target changes in appearance due to variations in viewpoint and lighting
conditions, changes in the pose of the target, partial occlusions, non-rigid
deformations of the target, and due to changes in the surrounding context and
background. In total there are 105 image pairs taken from 35 colour videos that
have previously been used as an object tracking benchmark
\citep{Wu_etal13b}. This dataset\footnote{\label{footnote-BBS}
  \url{http://people.csail.mit.edu/talidekel/Best-Buddies Similarity.html}.
  This dataset uses images that are 20 frames apart. A similar dataset with 25,
  50, or 100 frames between pairs of images was used to test the BBS algorithm in
  \citep{Oron_etal18}. However, this alternative dataset has not been made
  publically available.}
was originally prepared to evaluate the performance of the BBS template matching algorithm
\citep{Dekel_etal15}. Experimental procedures equivalent to those used in
\citet{Dekel_etal15} have also been used here. Specifically, the bounding-box in
the first image of each pair was used to define a template. The template
matching method was then used to calculate the similarity between this template
and every location in the second image of the same pair. The single location
which had the highest similarity was used as the predicted location of the
target. A bounding-box, the same size as the one in the first image, was defined
around this predicted location and the overlap between this and the bounding-box
provided in the ground-truth data was determined, and used as a measure of the
accuracy of the template matching. This bounding box overlap was calculated, in
the standard way, as the intersection over union (IoU) of the two bounding boxes.

\begin{figure*}[tbp]
  \begin{center}
    \includegraphics[width=0.22\textwidth,trim=0 0 350 0, clip]{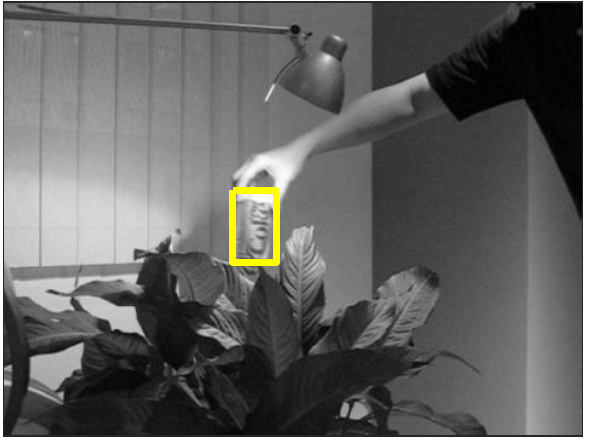}
    \includegraphics[width=0.22\textwidth,trim=0 0 350 0, clip]{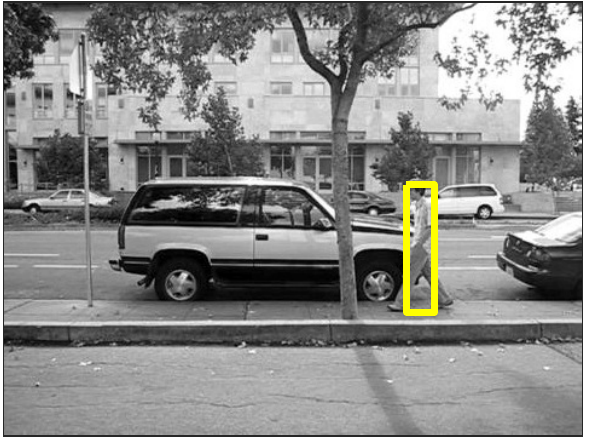}
    \includegraphics[width=0.22\textwidth,trim=0 0 350 0, clip]{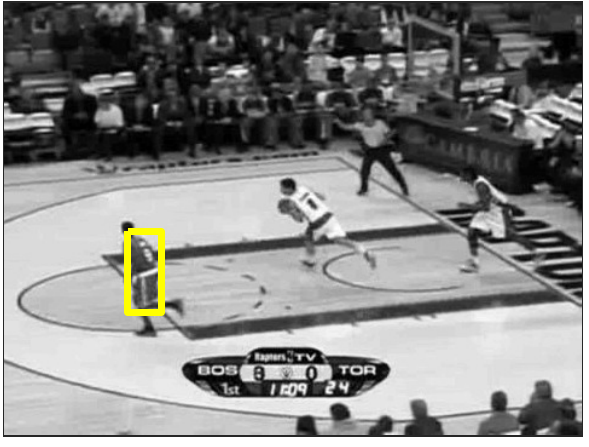}
    \includegraphics[width=0.22\textwidth,trim=0 0 350 0, clip]{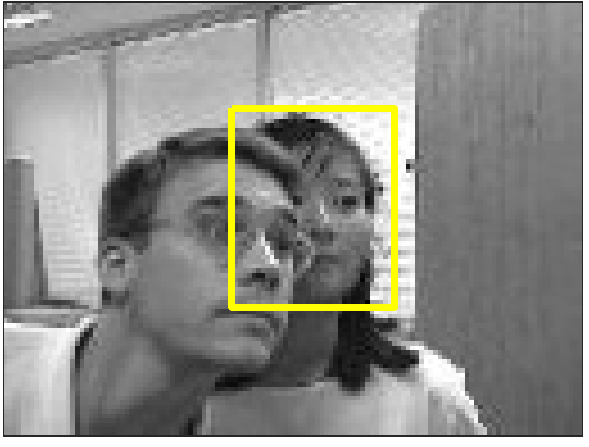}
    \rotatebox{90}{\hspace*{10mm}\textcolor{white}{{\small(1 tpl)}}}\hfill

    \includegraphics[width=0.22\textwidth,trim=175 0 175 0, clip]{bbs_data_pair10_DIM4}
    \includegraphics[width=0.22\textwidth,trim=175 0 175 0, clip]{bbs_data_pair22_DIM4}
    \includegraphics[width=0.22\textwidth,trim=175 0 175 0, clip]{bbs_data_pair81_DIM4}
    \includegraphics[width=0.22\textwidth,trim=175 0 175 0, clip]{bbs_data_pair88_DIM4}
    \rotatebox{90}{\hspace*{10mm}\textcolor{white}{{\small(1 tpl)}}}\hfill

    \includegraphics[width=0.22\textwidth,trim=350 0 0 0, clip]{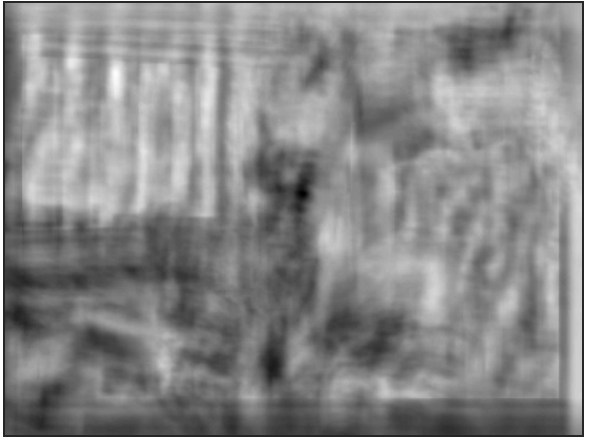}
    \includegraphics[width=0.22\textwidth,trim=350 0 0 0, clip]{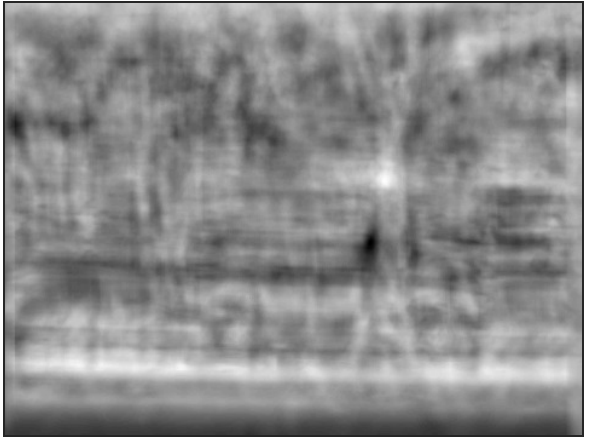}
    \includegraphics[width=0.22\textwidth,trim=350 0 0 0, clip]{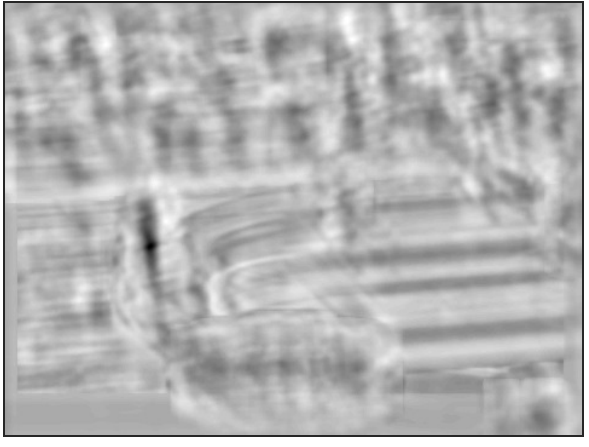}
    \includegraphics[width=0.22\textwidth,trim=350 0 0 0, clip]{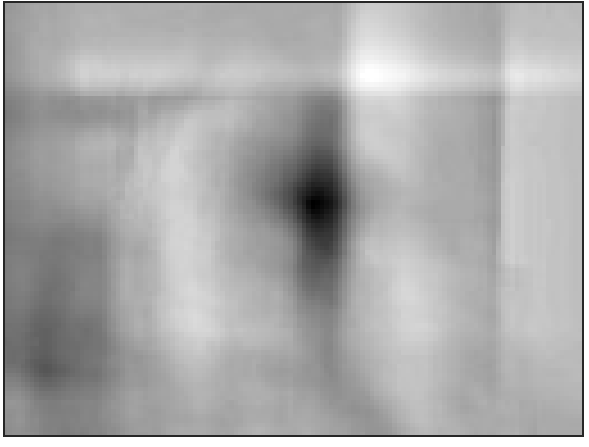}
    \rotatebox{90}{\hspace*{9.5mm}ZNCC\textcolor{white}{{\small(1 tpl)}}}\hfill

    \includegraphics[width=0.22\textwidth,trim=350 0 0 0, clip]{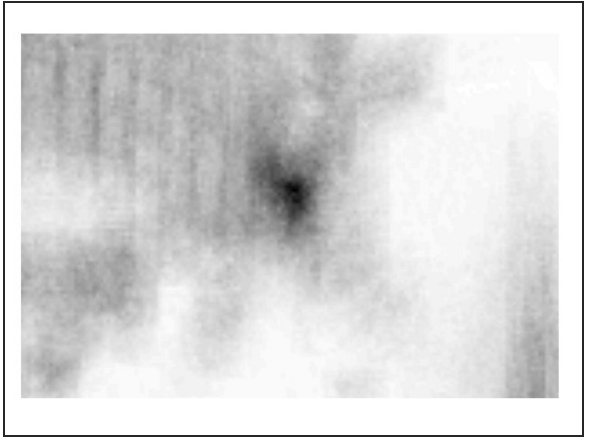}
    \includegraphics[width=0.2201\textwidth,trim=350 0 0 0, clip]{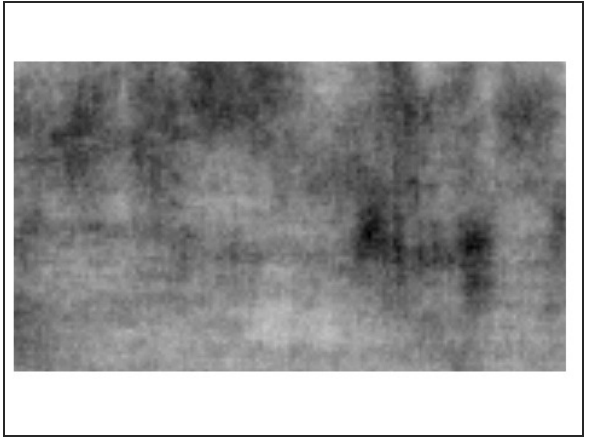}
    \includegraphics[width=0.2201\textwidth,trim=350 0 0 0, clip]{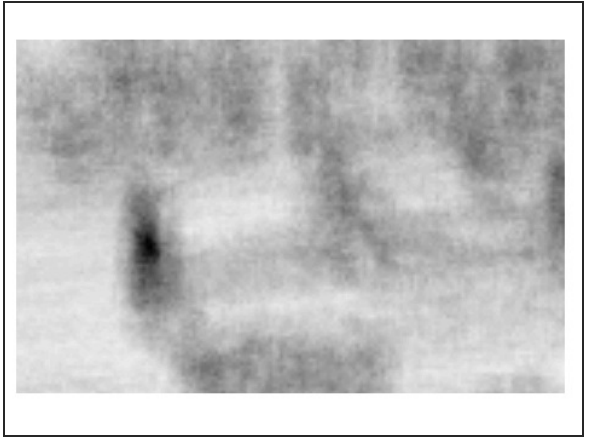}
    \includegraphics[width=0.22\textwidth,trim=350 0 0 0, clip]{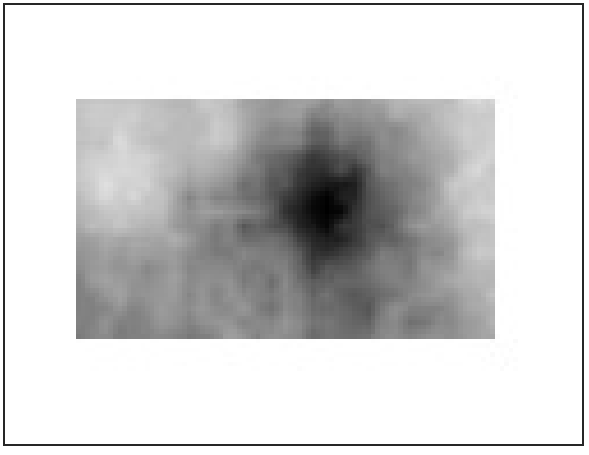}
    \rotatebox{90}{\hspace*{11mm}BBS\textcolor{white}{{\small(1 tpl)}}}\hfill

    \includegraphics[width=0.22\textwidth,trim=350 0 0 0, clip]{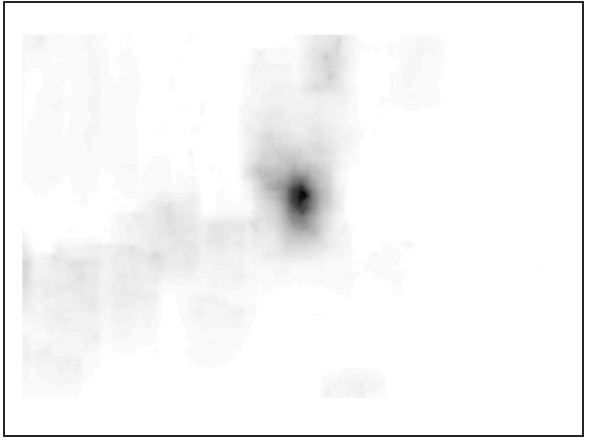}
    \includegraphics[width=0.22\textwidth,trim=350 0 0 0, clip]{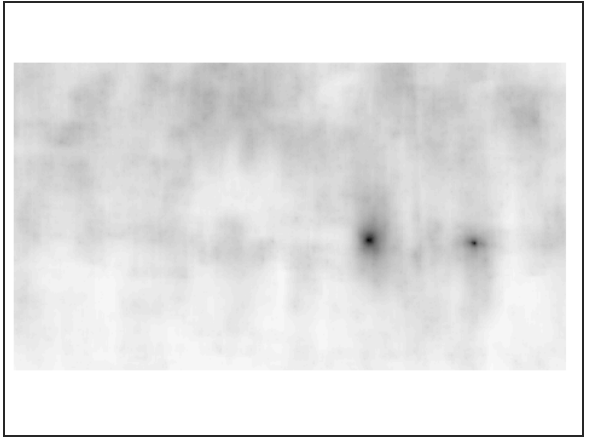}
    \includegraphics[width=0.22\textwidth,trim=350 0 0 0, clip]{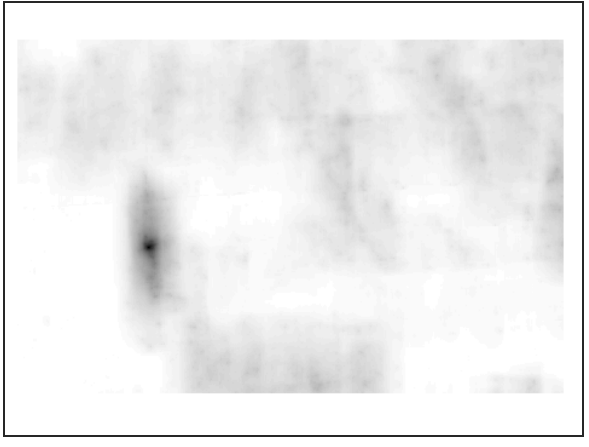}
    \includegraphics[width=0.22\textwidth,trim=350 0 0 0, clip]{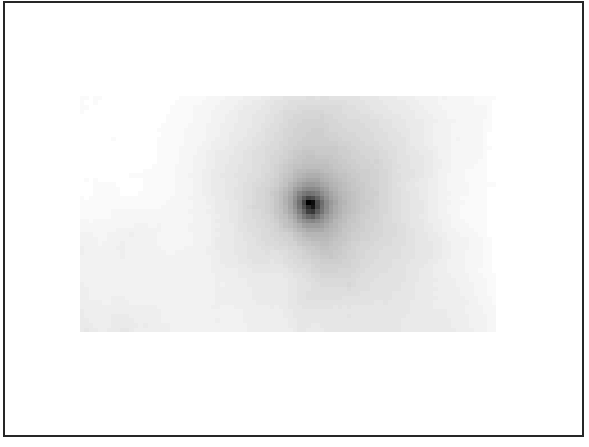}
    \rotatebox{90}{\hspace*{10mm}DDIS\textcolor{white}{{\small(1 tpl)}}}\hfill

    \includegraphics[width=0.22\textwidth,trim=350 0 0 0, clip]{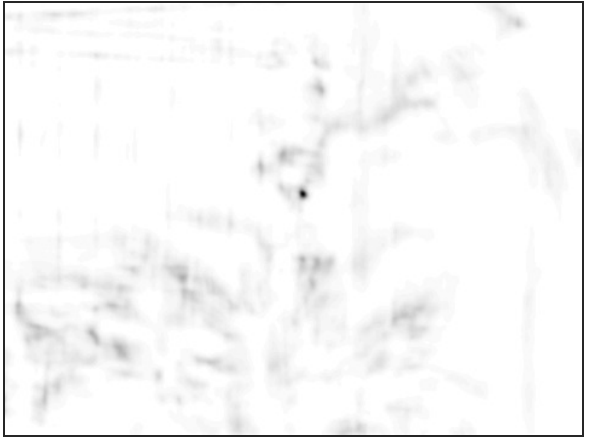}
    \includegraphics[width=0.22\textwidth,trim=350 0 0 0, clip]{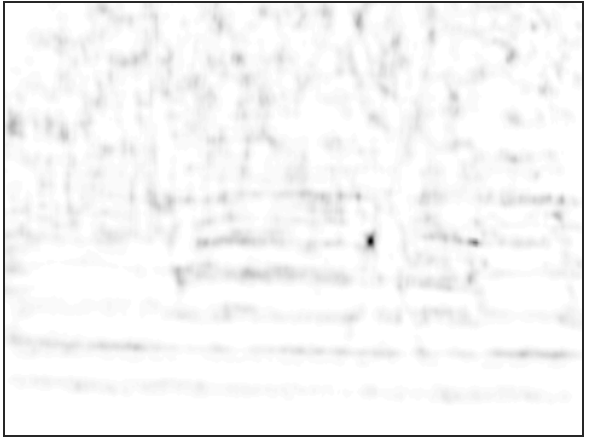}
    \includegraphics[width=0.22\textwidth,trim=350 0 0 0, clip]{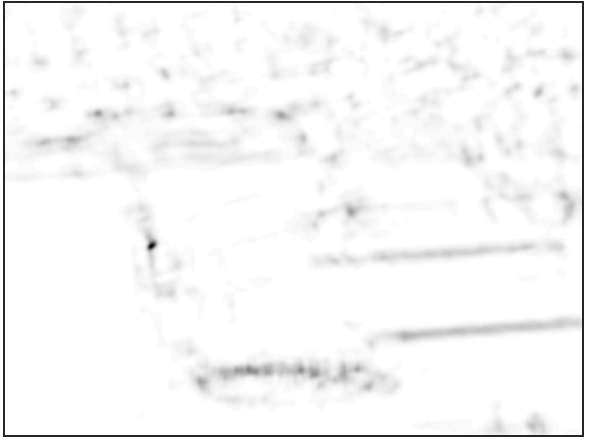}
    \includegraphics[width=0.22\textwidth,trim=350 0 0 0, clip]{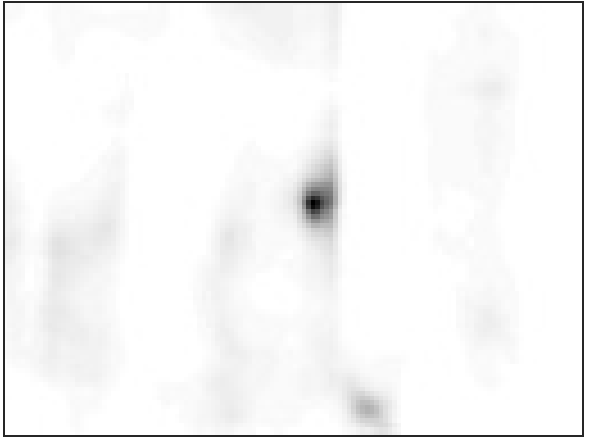}
    \rotatebox{90}{\hspace*{1.5mm}DIM {\small(1 template)}}\hfill

    \includegraphics[width=0.22\textwidth,trim=350 0 0 0, clip]{bbs_data_pair10_DIM4}
    \includegraphics[width=0.22\textwidth,trim=350 0 0 0, clip]{bbs_data_pair22_DIM4}
    \includegraphics[width=0.22\textwidth,trim=350 0 0 0, clip]{bbs_data_pair81_DIM4}
    \includegraphics[width=0.22\textwidth,trim=350 0 0 0, clip]{bbs_data_pair88_DIM4}
    \rotatebox{90}{\hspace*{11mm}DIM\textcolor{white}{{\small(1 tpl)}}}\hfill
    \caption{Example results for different algorithms when applied to the task
      of finding corresponding locations in 105 pairs of colour video frames.  Images
      in the first row show the target templates (outlined in yellow) in the
      initial frame of the video. Images in the second row show the location of
      the target identified by the DIM algorithm (outlined in cyan) and the
      location of the target defined by the ground-truth data (outlined in
      yellow) in a later frame of the same video. The third to seventh rows show
      the similarity of the target template to the second image as determined by
      (from row 3 to 7): ZNCC, BBS, DDIS, DIM with no additional templates, and
      DIM with up to four additional templates chosen by maximum
      correlation. Darker pixels correspond to stronger similarity. Note,
      matching was performed using colour templates and colour images, but for
      clarity the images are shown in grayscale in rows 1 and 2.}
    \label{fig-bss_data_examples}
  \end{center}
\end{figure*}

Results for typical example image pairs are shown in
\autoref{fig-bss_data_examples}. The images on the top row are the first images
in each of four image pairs. The yellow box superimposed on each image shows the
target image patch. The second row shows the second image in each pair. Two
boxes are superimposed on these second images. The yellow box shows the location
corresponding to the target defined by the ground-truth data. The cyan box shows
the location of the target predicted by the DIM algorithm. The remaining rows
show the similarity between the template and the second image calculated by
several different methods. The strongest measures of similarity are represented
by the darkest pixels. The similarity array calculated by ZNCC
(\autoref{fig-bss_data_examples} third row) is dense. The similarity arrays
produced by BBS (\autoref{fig-bss_data_examples} fourth row), DDIS
(\autoref{fig-bss_data_examples} fifth row), and DIM
(\autoref{fig-bss_data_examples} sixth and seventh rows), become increasingly
sparse, and hence, the peaks become increasingly well-localised and easily
distinguishable from non-matching locations.

Two results are shown for the DIM algorithm. The first result
(\autoref{fig-bss_data_examples} sixth row) shows the similarity of the target
template to the second image when only the target template was used by DIM.  The
second result for the DIM algorithm (\autoref{fig-bss_data_examples} seventh row) shows the
similarity calculated when up to four additional templates, the same size as the
bounding-box defining the target, were also extracted from the first image and
used as templates for non-target locations by DIM. These additional templates
were extracted from around locations where the correlation between the target
template and the image was strongest, excluding locations where the additional
templates would overlap with each other or with the bounding-box defining the
target. The exact number of additional templates varied between different image
pairs, and in some cases over the 105 image pairs, there were zero additional
templates due to the target bounding-box being large compared to the size of the
image. For the particular examples shown in \autoref{fig-bss_data_examples} all
results were produced using four additional templates, except that for the
right-most image where two additional templates were used.  As can be seen by
comparing the sixth and seventh rows of \autoref{fig-bss_data_examples},
including additional, non-target, templates tends to increase the sparsity of
the similarity array produced by DIM for the target template. This increased
sparsity results from the increased competition when there are more templates
(see \autoref{sec-methods_matching}).

\begin{figure}[tbp]
\begin{center}
  \subfigure[]{\includegraphics[scale=0.4]{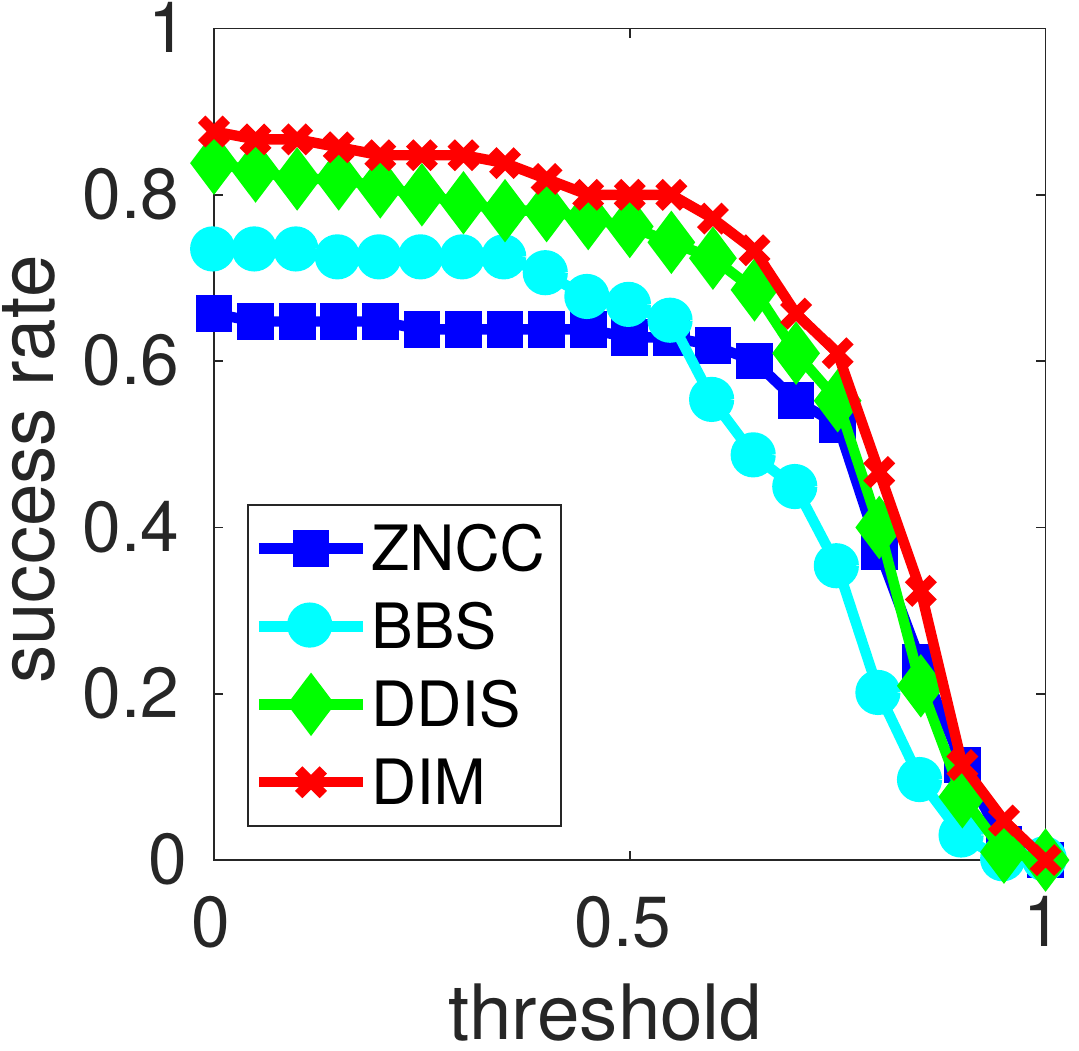}}
  \subfigure[]{\includegraphics[scale=0.4,trim=55 0 0 0, clip]{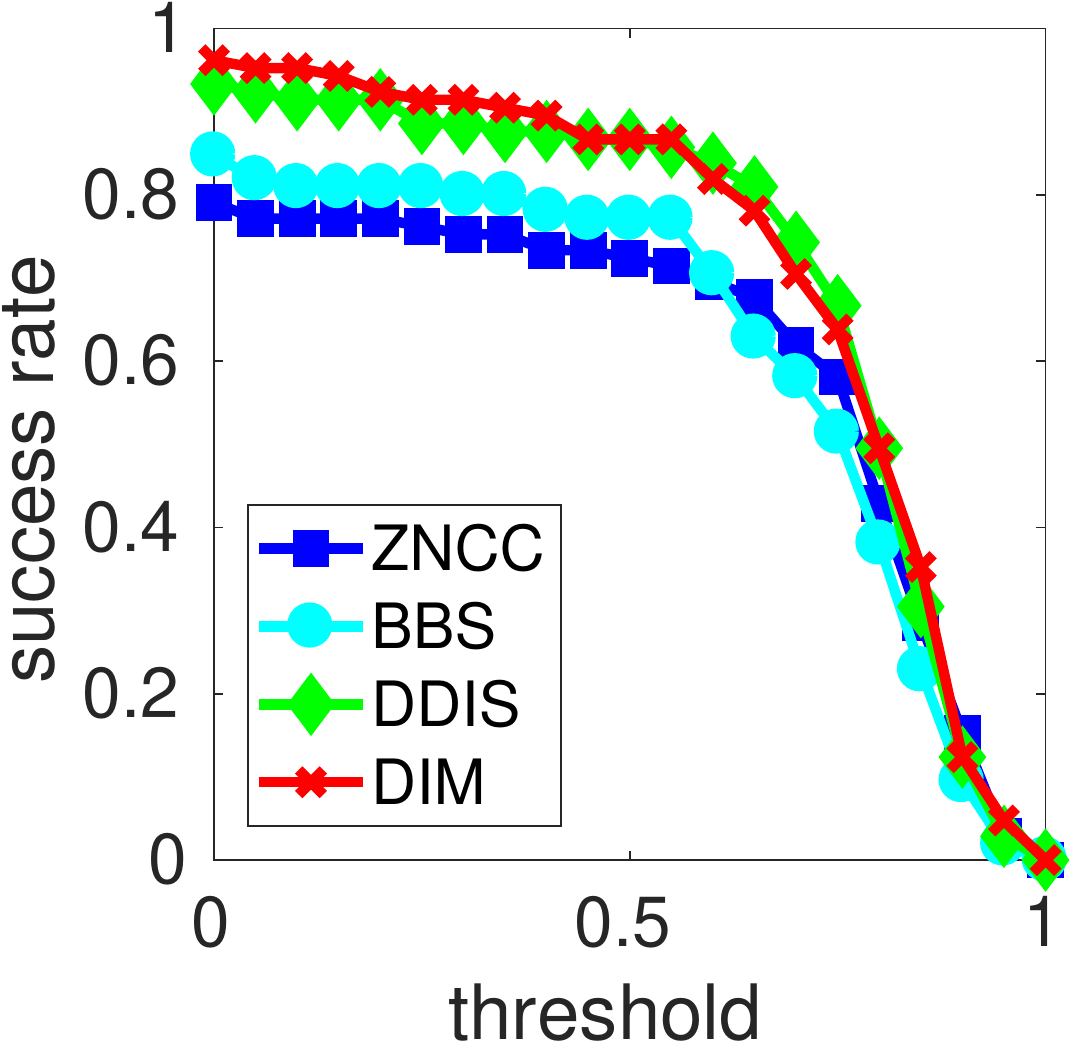}}
  \caption{The performance of different algorithms when applied to the task of finding
    corresponding locations in 105 pairs of colour video frames. Each curve shows the
    fraction of targets for which the overlap between the ground-truth and
    predicted bounding-boxes exceeded the threshold indicated on the x-axis. (a)
    Results when using the target location predicted by the maximum
    similarity. (b) Results when using the maximum overlap predicted by the
    seven highest similarity values. The results for DIM are produced using up
    to four additional templates chosen by maximum correlation.}
\label{fig-bbs_data_success}
\end{center}
\end{figure}

The results across all 105 image pairs are summarised in
\autoref{fig-bbs_data_success}(a). This graph shows the proportion of image
pairs for which the overlap between the ground-truth and the predicted
bounding-box (the IoU) exceeded a threshold for a range of different threshold
values. It can be seen that the success rate of DIM exceeds that of the other
methods at all thresholds.  Following the methods used in \citet{Dekel_etal15},
the overall accuracy of each method was summarised using the area under the
success curve (AUC). These quantitative results are shown in
\autoref{tab-bbs_data_AUC}, and compared to the published results for several
additional algorithms that have been evaluated on the same dataset.

\citet{Dekel_etal15} also assessed the accuracy of BBS by taking the largest
overlap across the seven locations with the highest similarity values. The same
analysis was done here by finding the seven largest peaks in the similarity
array for the target template, ignoring values that were not local maxima. These
results are presented in \autoref{fig-bbs_data_success}(b). It can be seen that
DIM produces similar results to DDIS, but significantly better performance than
both ZNCC and BBS over a wide range of threshold values.

\begin{table}[tbp]
\begin{center}
\dsoff
\begin{tabular}{ll} \hline
\tabind {\bf Algorithm} & {\bf AUC} \\ 
\hline

\emph{Baseline}\\
\tabind SSD                                               & 0.43 \citep{Dekel_etal15}  \\
\tabind NCC                                               & 0.47 \citep{Dekel_etal15}  \\
\tabind SAD                                               & 0.49 \citep{Dekel_etal15}  \\
\tabind ZNCC                                              & 0.54\\ 

\emph{State-of-the-art}\\
\tabind BBS \citep{Dekel_etal15,Oron_etal18}              & 0.55 \citep{Dekel_etal15}\\ 
\tabind CNN (2ch-deep) \citep{ZagoruykoKomodakis15,ZagoruykoKomodakis17}       & 0.59 \citep{Kim_etal17}\\
\tabind CoTM \citep{Kat_etal18}                           & 0.62$^*$ \citep{Kat_etal18} \\ 
\tabind CNN (SADCFE) \citep{Kim_etal17}                   & 0.63 \citep{Kim_etal17}\\
\tabind CNN (2ch-2stream) \citep{ZagoruykoKomodakis15,ZagoruykoKomodakis17}    & 0.63 \citep{Kim_etal17}\\
\tabind DDIS \citep{Talmi_etal17}                         & 0.64 \citep{Kat_etal18}  \\ 

\emph{Proposed}\\
\tabind DIM (1 template)                                  & 0.58 \\ 
\tabind DIM (1 to 4 templates)                            & {\bf 0.69} \\ 
\hline
\end{tabular}
\caption{Quantitative comparison of results for different algorithms when
  applied to the task of finding corresponding locations in 105 pairs of colour video
  frames. Results are given in terms of the area under the success curve
  (AUC). $^*$We failed to reproduce the result for CoTM given in
  \citep{Kat_etal18}: using the code released by the
  authors\protect\footnotemark\xspace to perform the matching, together with our
  code to run the benchmark, produced an AUC of 0.54. Using the same code for
  running the benchmark and the code supplied by the original authors it was
  possible to reproduce the published results for both BBS and DDIS.}
\label{tab-bbs_data_AUC}
\end{center}
\end{table}
\footnotetext{\url{http://www.eng.tau.ac.il/~avidan/}}

From \autoref{tab-bbs_data_AUC} it can be seen that the performance of DIM is
strongly effected by the inclusion of additional templates extracted from other
parts of the first image. This is to be expected, as the proposed method
performs explaining away, and the accuracy of this inference process will
increase when additional templates compete with the target template such that
regions of the image that should have low similarity with the target template
are explained-away by the additional templates. Without any additional templates
the performance of DIM is more similar to ZNCC, this is also to be expected as both
methods are comparing the relative intensity values in the template and each
patch of image.

\begin{figure}[tbp]
\begin{center}
  \subfigure[]{\includegraphics[scale=0.4]{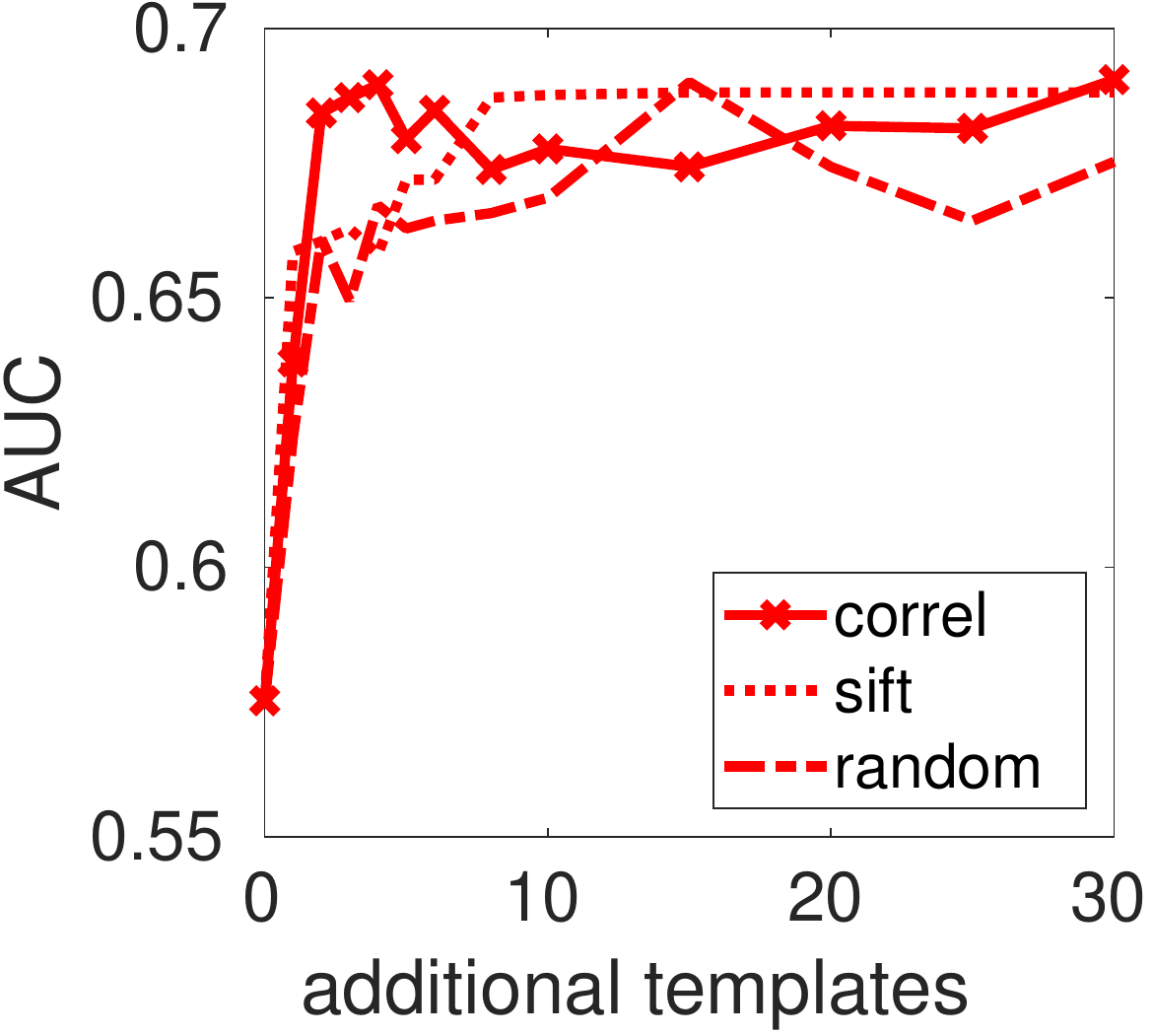}\label{fig-bbs_data_num_templates}}
  \subfigure[]{\includegraphics[scale=0.4,trim=70 0 0 0, clip]{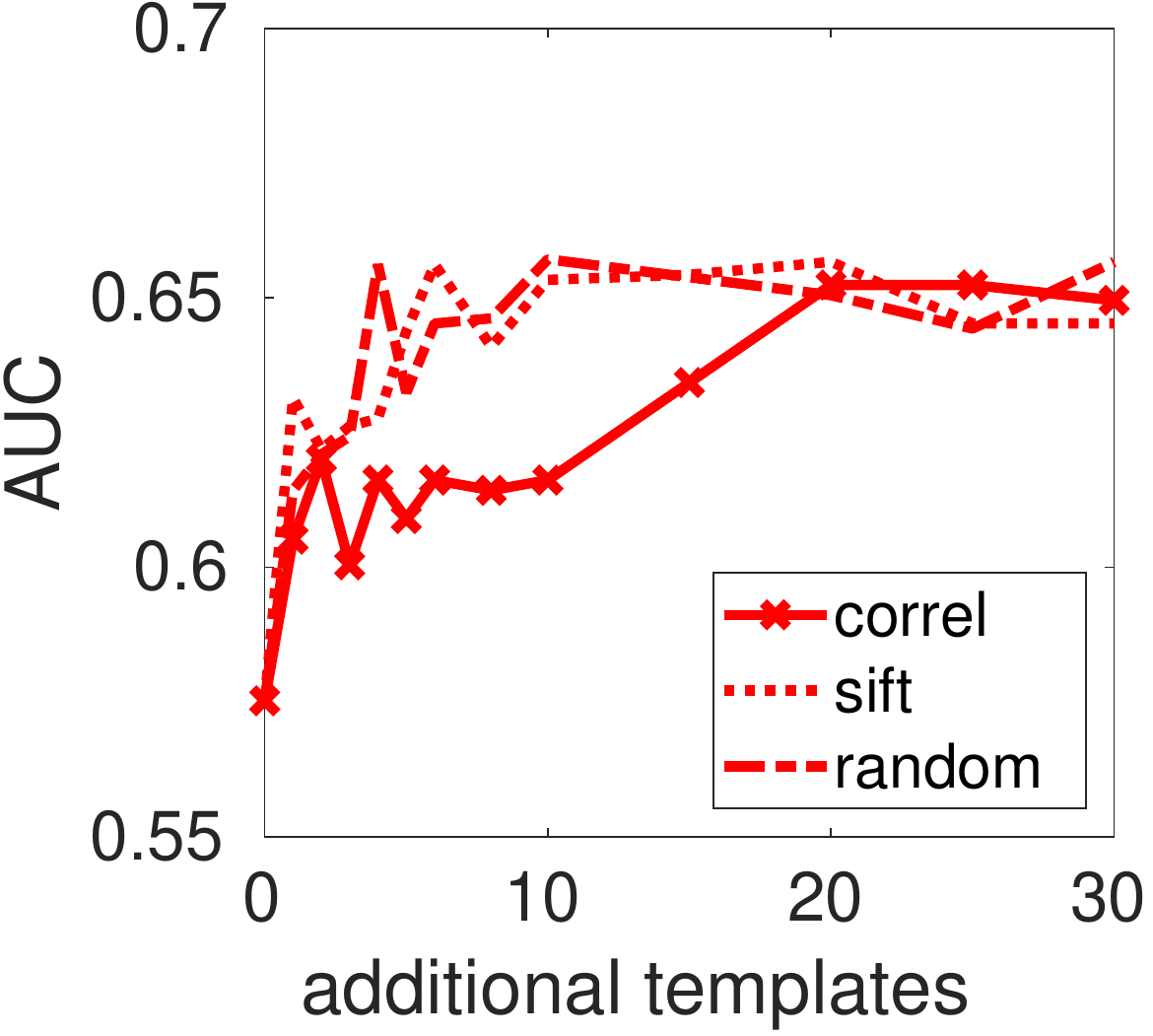}\label{fig-bbs_data_num_templates_background}}
  \caption{The effect of the maximum number of additional templates, and their
    selection method, on the performance of the proposed method, DIM, when
    applied to finding corresponding locations in 105 pairs of colour video
    frames. Results are shown when the additional templates were selected from
    (a) the first image in each pair, and (b) an unrelated image.}
  \label{fig-bbs_data_num_templates_both}
\end{center}
\end{figure}

A number of different methods of choosing the additional, non-target, templates
used by DIM were investigated. \Autoref{fig-bbs_data_num_templates} shows the
AUC obtained for different maximum numbers of additional templates, when these
templates were selected from locations where the correlation between the target
template and the image was strongest, from locations selected by the SIFT
interest point detector, and locations chosen at random.  In each case,
additional templates were chosen such that they did not overlap with each other
or with the bounding-box defining the target so as to ensure a diversity of
additional templates. The exact number of additional templates varied between
different image pairs, depending on how many non-overlapping regions, equal in
size to the target template, could fit within the first image in each pair.
It can be seen that the performance of DIM increased as the number of non-target
templates increased, and that for a large number of additional templates the AUC
plateaued between 0.66 and 0.69 regardless of how the additional templates were
selected. Furthermore, for all three methods of selecting additional templates
the performance of DIM exceeded that of the current state-of-the-art method,
DDIS, when two or more additional templates were used. It can also be observed
from \autoref{fig-bbs_data_num_templates} that the initial increase in performance
with the number of additional templates was faster when the non-target templates
were chosen by maximum correlation. In other words, the best performance was
achieved with fewer additional templates if those additional templates were
chosen so that they were the regions of the first image that were most similar to,
and hence most easily mistaken for, the target region.

One concern is that the benefits of performing explaining away will disappear
when the target appears in a completely different context. In other words, if
the background of the first image is different from that for the second image,
then additional, non-target, templates extracted from the first image will be
ineffective at competing with the target template when they are matched to the
second image. To explore this issue, the experiment described in the preceding
paragraph was repeated, but the additional templates were taken from an
unrelated image (the first ``Leuven'' image from the Oxford VGG Affine Covariant
Features Dataset \citep{MikolajczykSchmid05,Mikolajczyk_etal05}, see
\autoref{sec-correspondence_vgg}). The results of this experiment are shown in
\autoref{fig-bbs_data_num_templates_background}.  As expected, there was a
deterioration in performance. However, as long as sufficient ($\ge20$)
additional templates were used then the performance of DIM (an AUC of $\ge64$) was
still as good as, or better than, all other methods that have been applied to
this benchmark (see \autoref{tab-bbs_data_AUC}).

\subsection{Correspondence using the Oxford VGG Affine Covariant Features Dataset}
\label{sec-correspondence_vgg}

This section describes an experiment similar to that in
\autoref{sec-correspondence_bbs}, but using the images from the Oxford VGG
Affine Covariant Features
Benchmark\footnote{\url{http://www.robots.ox.ac.uk/~vgg/research/affine/}}
\citep{MikolajczykSchmid05,Mikolajczyk_etal05}. This dataset has been widely
used to test the ability of interest point detectors to locate corresponding
points in two images.  The dataset consists of eight image sequences (seven
colour and one grayscale). Each sequence consists of six images of the same
scene.  These images differ in viewpoint (resulting in changes in perspective,
orientation, and scale), illumination/exposure, blur/de-focus and JPEG
compression. The ground-truth correspondences are defined in terms of
homographies (plane projective transformations) which relate any location in the
first image of each sequence to its corresponding location in the remaining five
images in the same sequence. In this experiment, templates were extracted from
the first image in each sequence and the best matching locations were found in
each of the remaining five images in the same sequence.

Images were scaled to half their original size to reduce the time taken to
perform this experiment.  From the first image in each sequence templates were
extracted (as described in \autoref{sec-methods_preproc}). These templates were
extracted from around keypoints identified using the Harris corner detector. A
keypoint detector was used to identify suitable locations for matching in order
to exclude locations that no algorithm could be expected to match such as
regions of uniform colour and luminance. The results were not dependent on the
particular keypoint detector used. From the first image in each sequence 25
keypoints were chosen after excluding those for which: 1) the bounding box
defining the extent of the template was not entirely within the image; 2) the
bounding box around the corresponding location in the query image was not
entirely within the image; 3) the Manhattan distance between the keypoints was
less than 24 pixels, or less than the the size of the bounding box defining the
extent of the template (which ever distance was smaller). These criteria for
rejecting keypoints ensured that the templates did not fall off the edge of
either image in each pair (criteria 1 and 2), and increased the diversity of
image features that were being matched (criteria 3).

For the DIM algorithm, no additional templates were used as the 25 templates
extracted from the first image competed with each other, and hence, for each
template the remaining 24 templates effectively acted as additional templates
representing non-target image features.
  
\begin{figure*}[tbp]
  \begin{center}
    \includegraphics[scale=0.4,trim=0 20 110 0, clip]{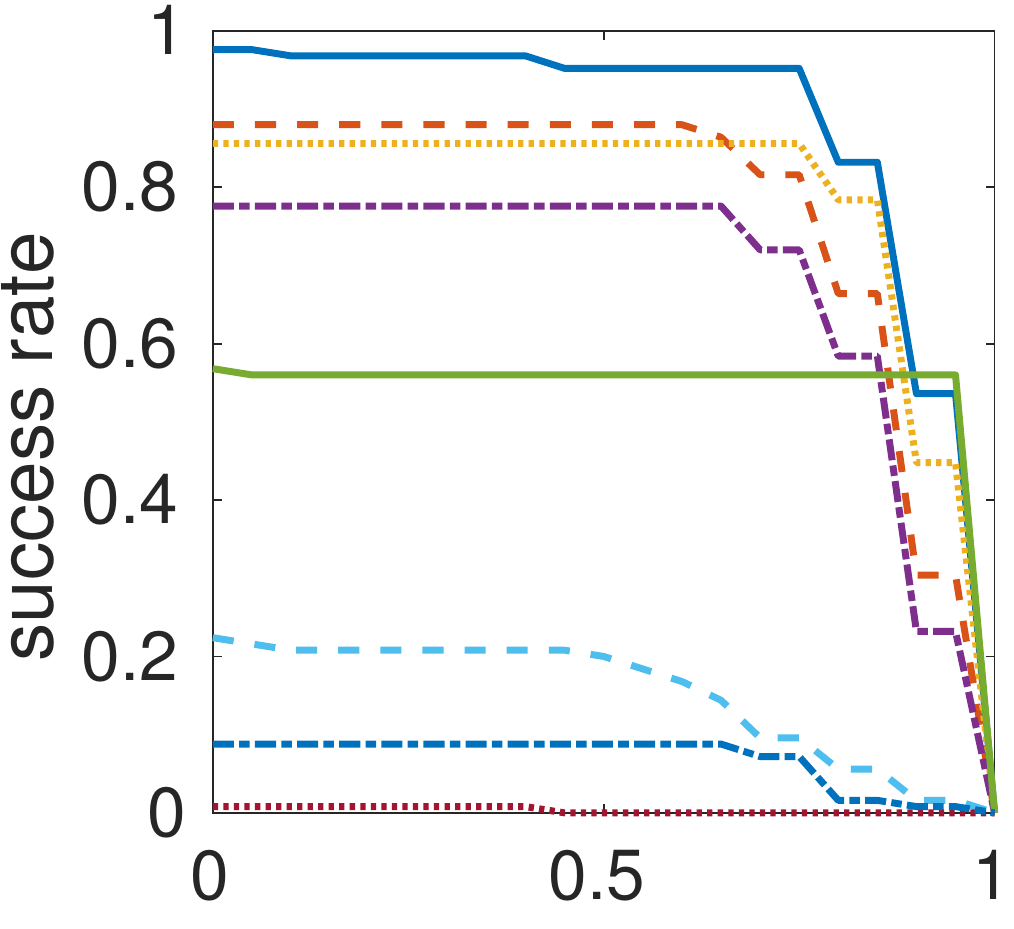}
    \includegraphics[scale=0.4,trim=20 20 110 0, clip]{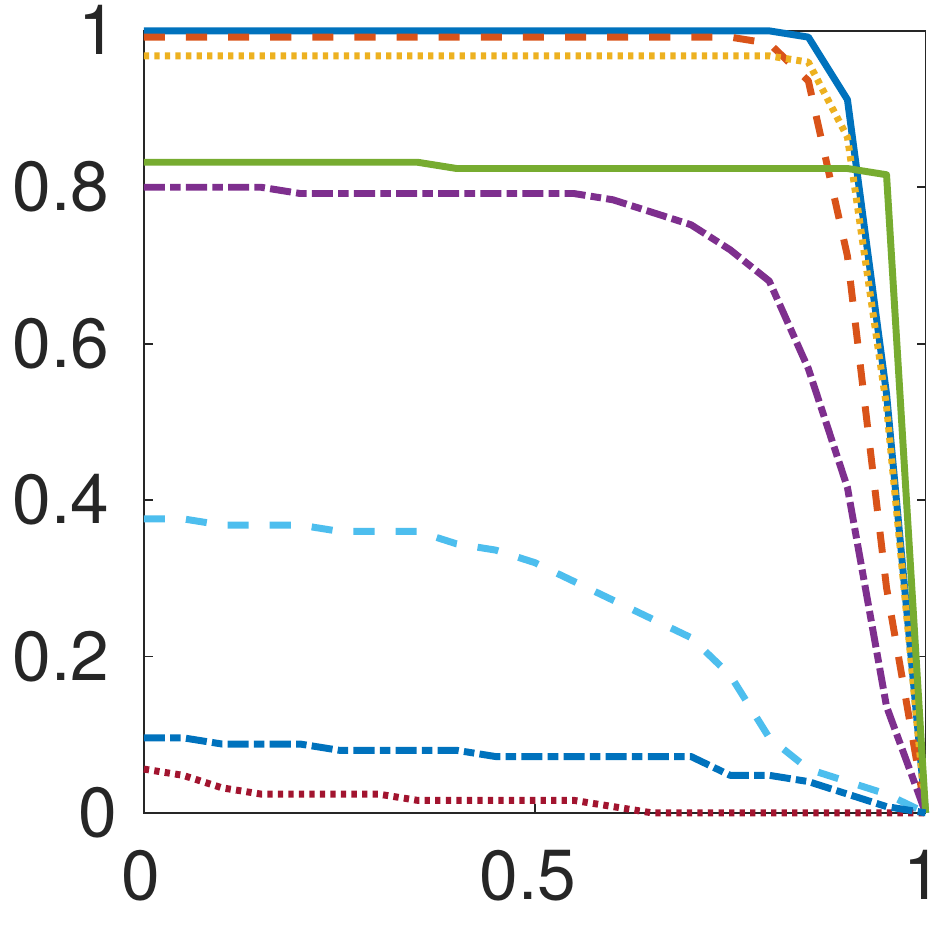}
    \includegraphics[scale=0.4,trim=20 20 110 0, clip]{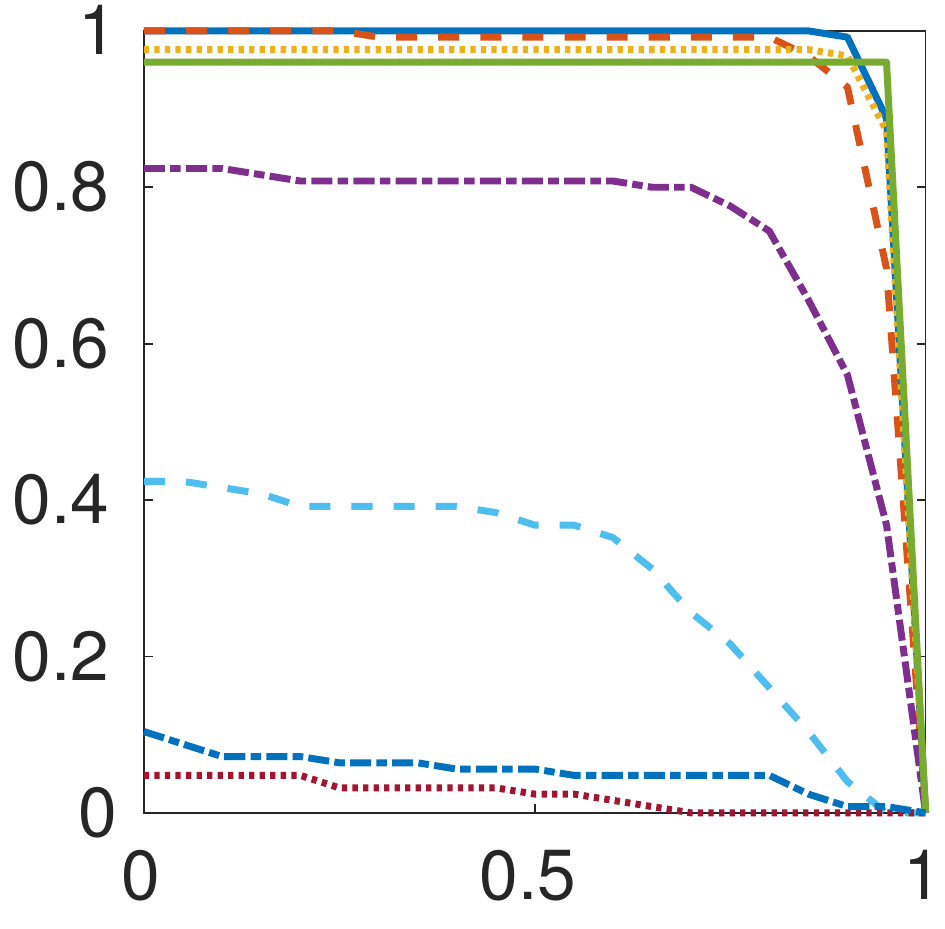}
    \rotatebox{90}{\hspace*{17mm}ZNCC}

    \includegraphics[scale=0.4,trim=0 20 110 0, clip]{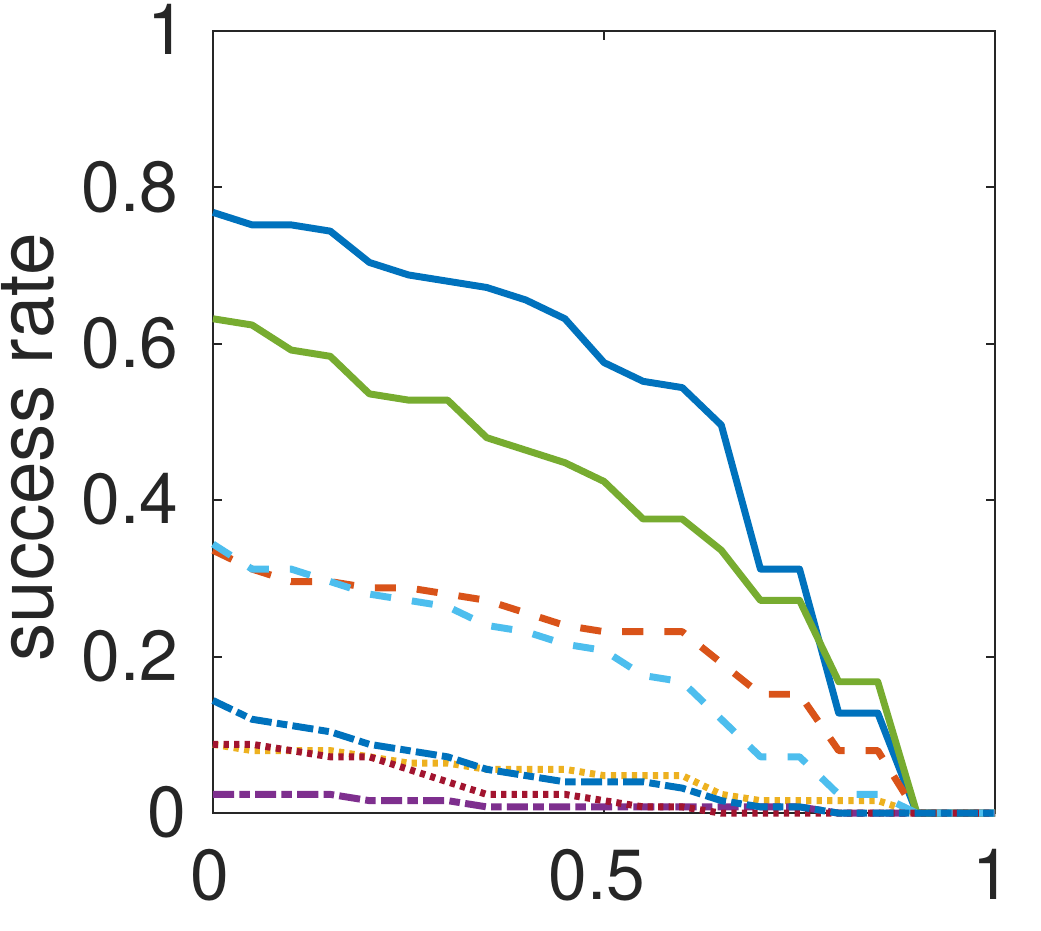}
    \includegraphics[scale=0.4,trim=20 20 110 0, clip]{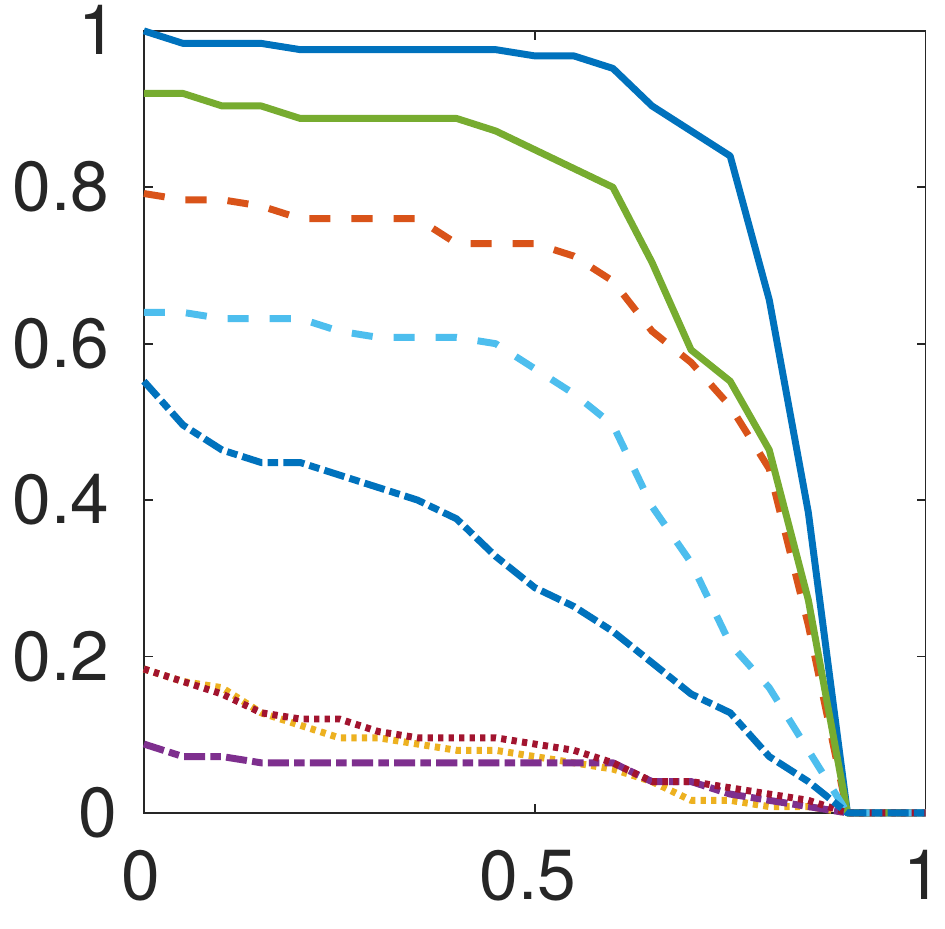}
    \includegraphics[scale=0.4,trim=20 20 110 0, clip]{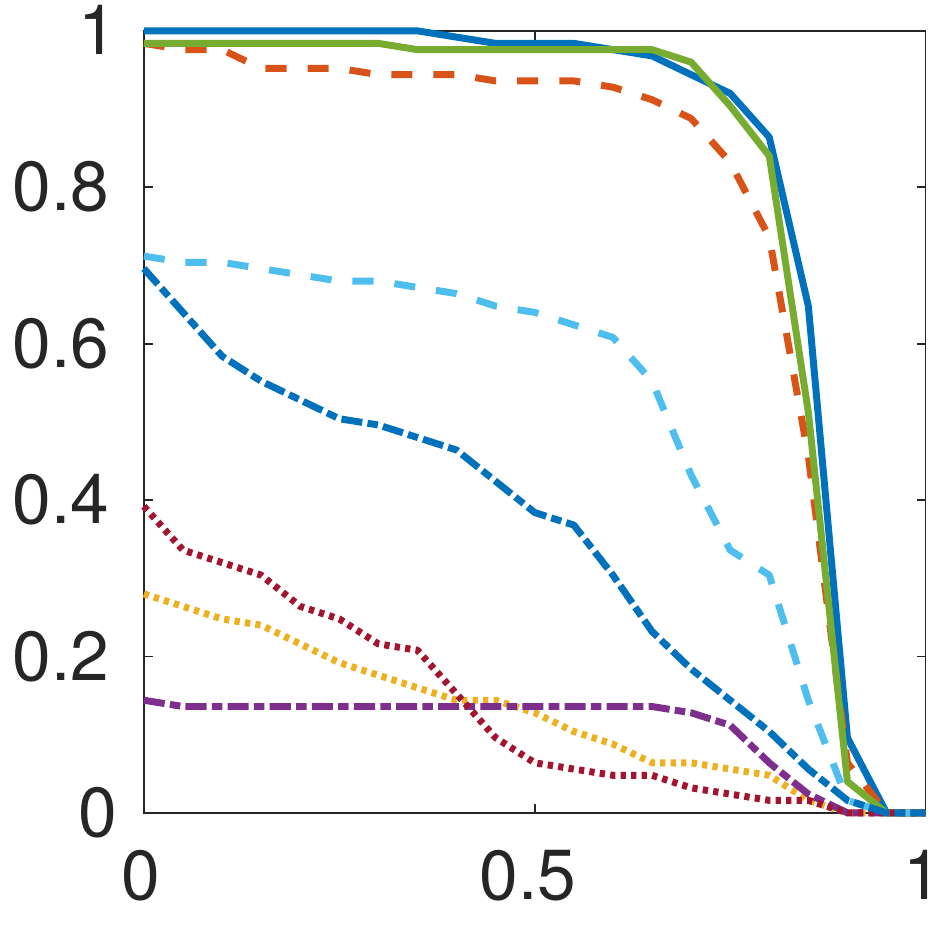}
    \rotatebox{90}{\hspace*{17mm}BBS}

    \includegraphics[scale=0.4,trim=0 20 110 0, clip]{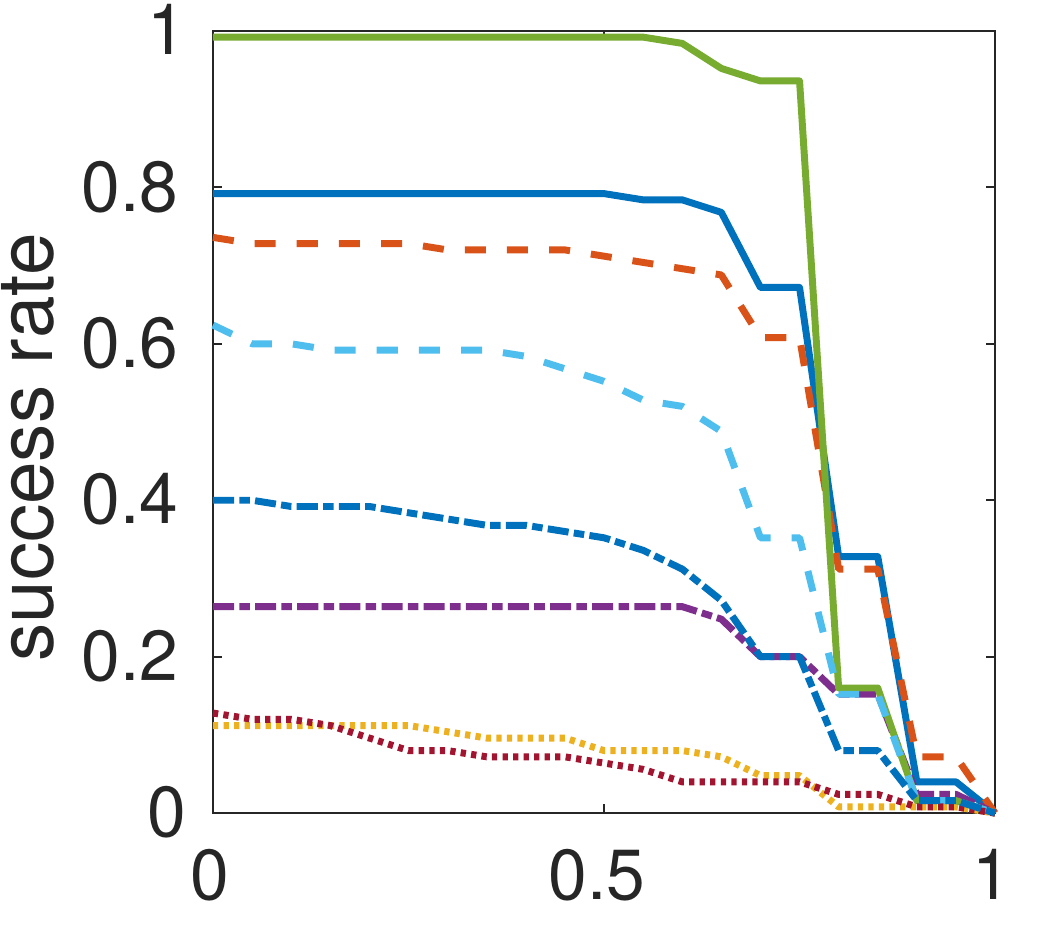}
    \includegraphics[scale=0.4,trim=20 20 110 0, clip]{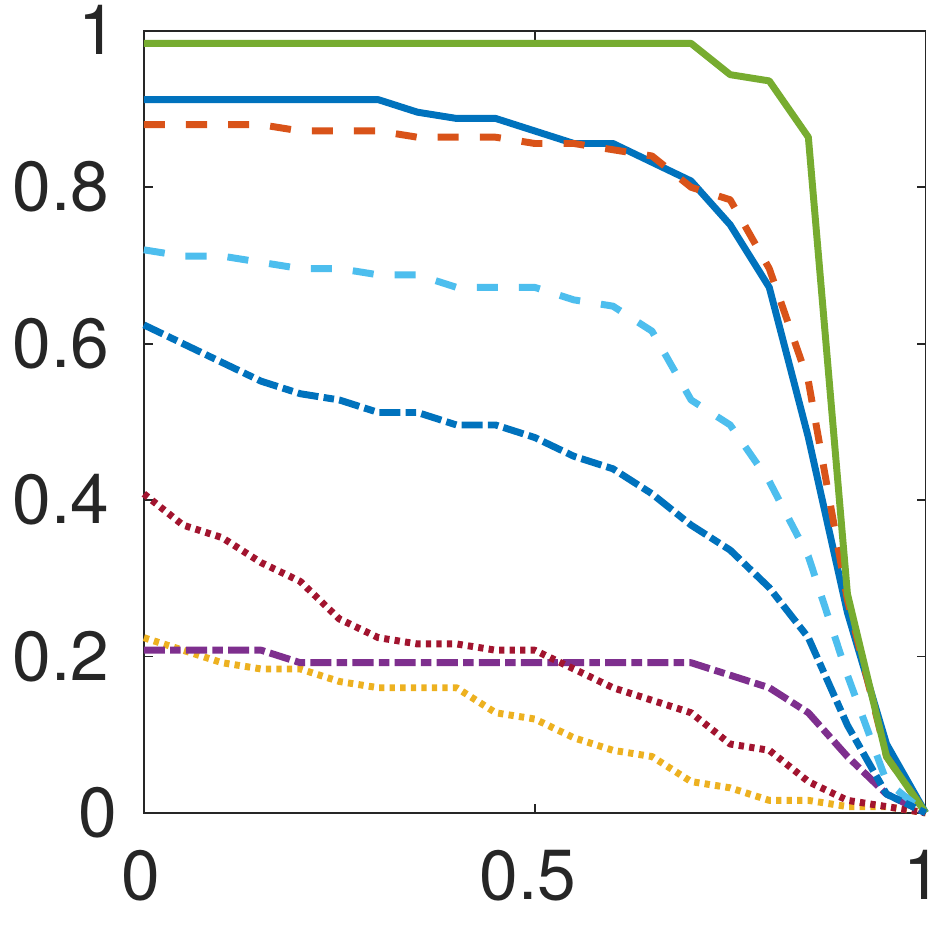}
    \includegraphics[scale=0.4,trim=20 20 110 0, clip]{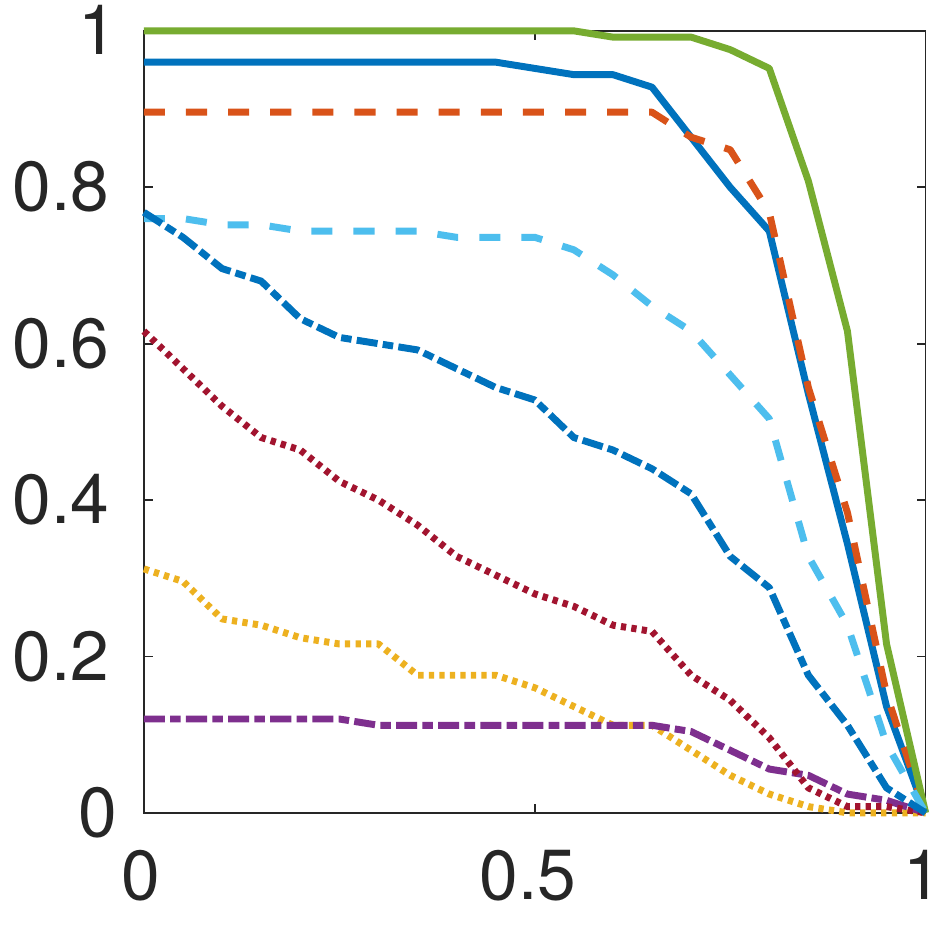}
    \rotatebox{90}{\hspace*{17mm}DDIS}

    \includegraphics[scale=0.4,trim=0 0 117 0, clip]{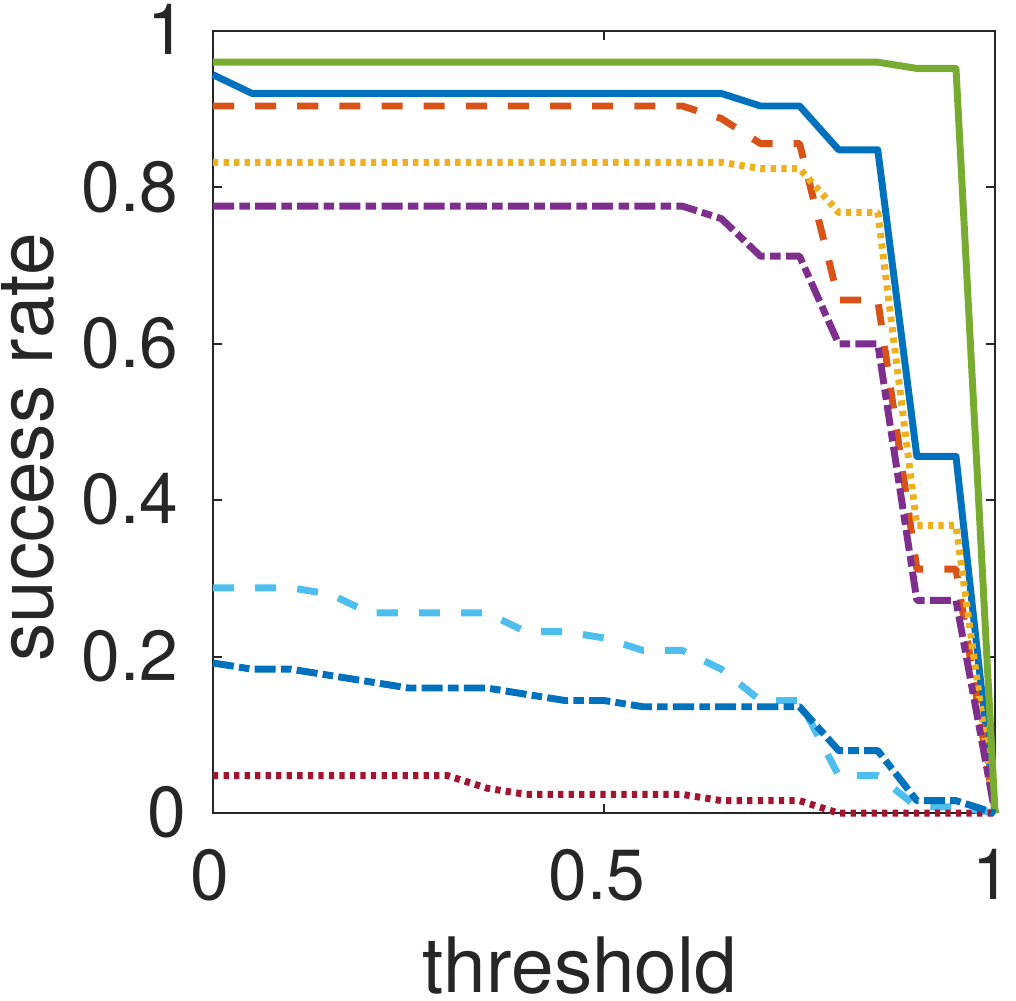}
    \includegraphics[scale=0.4,trim=20 0 117 0, clip]{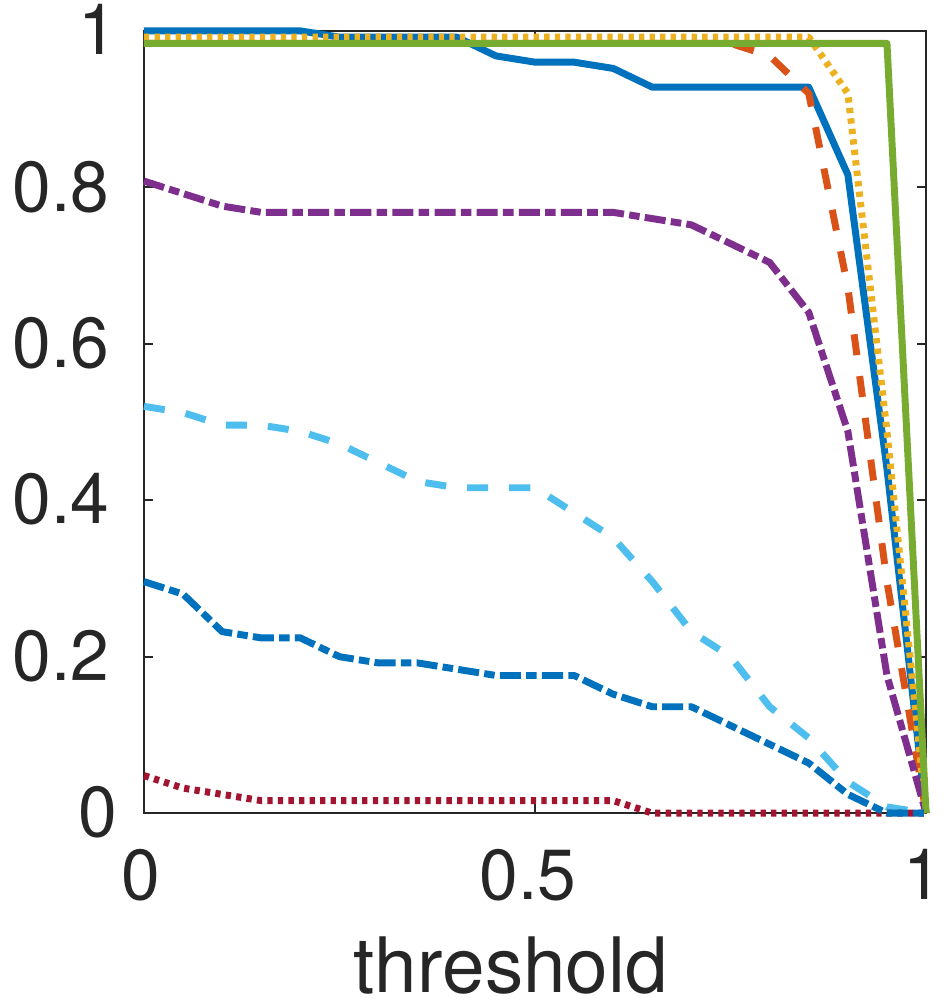}
    \includegraphics[scale=0.4,trim=20 0 117 0, clip]{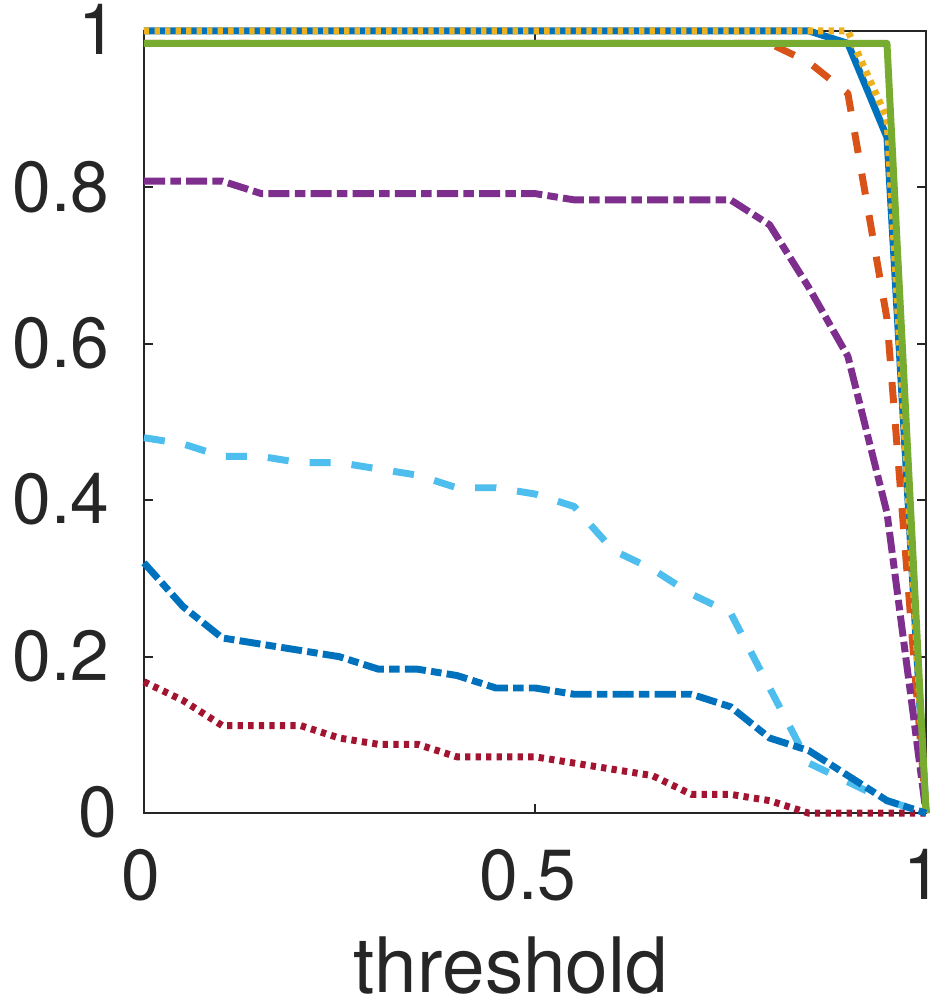}
    \rotatebox{90}{\hspace*{20mm}DIM}\\[1.5mm]

    \includegraphics[width=0.8\textwidth,trim=13 0 0 0, clip]{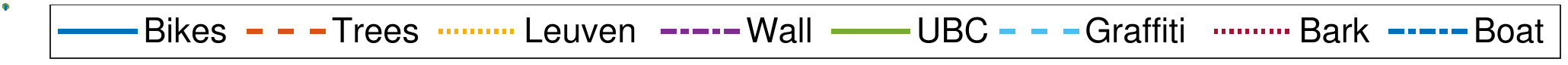} \\
    \caption{The performance of different algorithms when applied to the task of
      finding corresponding locations across image sequences from the Oxford
      VGG affine covariant features dataset (at half size with 25 templates per
      image pair). Each curve shows the fraction of targets for which the
      overlap between the ground-truth and predicted bounding-boxes exceeded the
      threshold indicated on the x-axis. Results for different image sequences
      are shown using different line styles, as indicated in the key.  Each row
      shows results for a different algorithm (from top to bottom): ZNCC, BBS,
      DDIS, and DIM. Each column shows results for a different template size
      (from left to right): 17-by-17, 33-by-33, and 49-by-49 pixels.}
    \label{fig-vgg_data_success}
  \end{center}
\end{figure*}

The success curves produced by each algorithm for each sequence are shown in
\autoref{fig-vgg_data_success}, for three different sizes of templates. For all
four methods the results generally improved as the template size increased.
Differences in appearance between images in the Bikes and Trees sequences are
primarily due to changes in image blur. It can be seen from
\autoref{fig-vgg_data_success} that all four algorithms produced some of their
strongest performances when matching locations on these images. The exception was
BBS which produced poor performance on the Trees sequence when the template size
was small. This is likely to be due to metric used by BBS being insufficiently
discriminatory to distinguish distinct locations in the leaves of the
trees. Images in the Leuven sequence vary primarily in terms of illumination and
exposure. ZNCC and DIM accurately matched points across these images. In contrast,
DDIS and BBS showed very little tolerance to illumination changes.  Differences in
appearance between images in the Wall and Graffiti sequences are primarily due
to changes in viewpoint. On the Wall sequence, ZNCC and DIM produced good
performance, while DDIS and BBS produced poor performance. In contrast, on the
Graffiti sequence DDIS produced the best performance of the four methods. These
differences are likely due to the Graffiti sequence having more distinctive
image regions, while the Wall images contain many similar looking locations as
it is a brick texture. Images in the UBC sequence differ in their JPEG quality. It can
be seen from \autoref{fig-vgg_data_success} that all four algorithms produced
some of their strongest performances when matching locations on these images,
except ZNCC and BBS for the small template sizes where the performance was
mediocre. Differences in appearance between images in the Bark and Boat
sequences are primarily caused by changes in scale and in-plane rotation. Both
these sequences were among the most challenging for all four methods. However,
the performance for both DDIS and BBS improved as the template size increased.
 
\begin{table}[tbp]
\begin{center}
\dsoff
\begin{tabular}{lllll} \hline
 {\bf Algorithm}                     &  & {\bf AUC}     & & \\   
\multicolumn{2}{r}{patch size (pixels):}& 17-by-17      & 33-by-33      & 49-by-49 \\
\hline
 ZNCC                                &  & 0.4996        & 0.5937        & 0.6314 \\ 
 BBS \citep{Dekel_etal15,Oron_etal18}&  & 0.1782        & 0.3834        & 0.4747 \\ 
 DDIS \citep{Talmi_etal17}           &  & 0.3952        & 0.4905        & 0.5334 \\
 DIM (25 templates)                  &  &{\bf 0.5591}   &{\bf 0.6308}   &{\bf 0.6569}\\
\hline
\end{tabular}
\caption{Quantitative comparison of results for different algorithms applied to
  the task of finding corresponding locations across images in the Oxford
  VGG affine covariant features dataset (at half size).  Results are given in
  terms of the area under the success curve (AUC) for all 1000 template-image
  comparisons across all eight sequences in the dataset (25 templates per image
  pair).}
\label{tab-vgg_data_AUC}
\end{center}
\end{table}

The overall accuracy of each method was summarised using the area under the
success curve (AUC) for all 1000 template matches performed (25 templates
matched to 5 images in each of 8 sequences). As the same number of template
matches were performed for each sequence, this is equivalent to the AUC for the
average of the individual success curves for the eight sequences shown in
\autoref{fig-vgg_data_success}. These quantitative results are shown in
\autoref{tab-vgg_data_AUC}. It can be seen that the proposed method, DIM,
significantly out-performs the other methods on this task. Surprisingly, both
BBS and DDIS are less accurate than the baseline method, ZNCC.

\subsection{Template Matching using the Oxford VGG Affine Covariant Features Dataset}
\label{sec-template_matching_vgg}

In the preceding two sections template matching algorithms have been evaluated
using correspondence tasks. In such tasks it is assumed that the target always
appears in the query image.  However, in many real-world applications such an
assumption is not appropriate, as it is not known if the searched-for image
feature appears in the query image. In such applications it is, therefore, not
appropriate to select the single location with the highest similarity to the
template as the matching location. Instead, it is necessary to apply a threshold
to the similarity values to distinguish locations where the template matches the
image from those where it does not. To avoid counting multiple matches within a
small neighbourhood, it is also typically the case that the similarity must be a
local maxima as well as exceeding the global threshold.

To evaluate the ability of the proposed method to perform template matching
under these conditions an experiment was performed using the colour images (\ie
excluding the Boat sequence) from the Oxford VGG Affine Covariant Features
Benchmark \citep{MikolajczykSchmid05,Mikolajczyk_etal05}. Only the colour images
were used as in this experiment templates extracted from one sequence were
matched to images from the other sequences: it was therefore necessary to have
all templates and query images either in colour or grayscale.
Images were scaled to one-half their original size to reduce the time taken
to perform this experiment. From the first image in each sequence 10 templates
were extracted from around keypoints identified using the Harris corner detector
and using the same criteria as described in \autoref{sec-correspondence_vgg}. A
total of 70 templates were thus defined (10 for each of the 7 colour
sequences). All these 70 templates were matched to each of the 5 query images in each
sequence (\ie to 35 colour images): a total of 2450 template-image comparisons in all.

For every location in the similarity array that was both a local maxima and
exceeded a global threshold a bounding box, the same size as the template, was
defined in the query image. These locations found by the template matching
method were compared to the true location that matched that template in the
query image: if the template came from the first image in the same sequence, the
comparison was with a bounding box (the same size as the template) defined
around the transformed location of the keypoint from around which the template
had been extracted; if the template came from a different sequence then there
was no matching bounding box in the query image. If the two bounding boxes (one
predicted by template matching and one from the ground truth data), had an
overlap (IoU) of at least 0.5 this was counted as a true-positive. A
ground-truth bounding box not predicted by the template matching process was
counted as a false-negative, while matches found by the template matching
algorithm that did not correspond to the ground-truth bounding box (or multiple
matches to the same ground-truth) were counted as false-positives.

\begin{figure*}[tbp]
\begin{center}
  \subfigure[]{\includegraphics[scale=0.4,trim=0 0 0 0, clip]{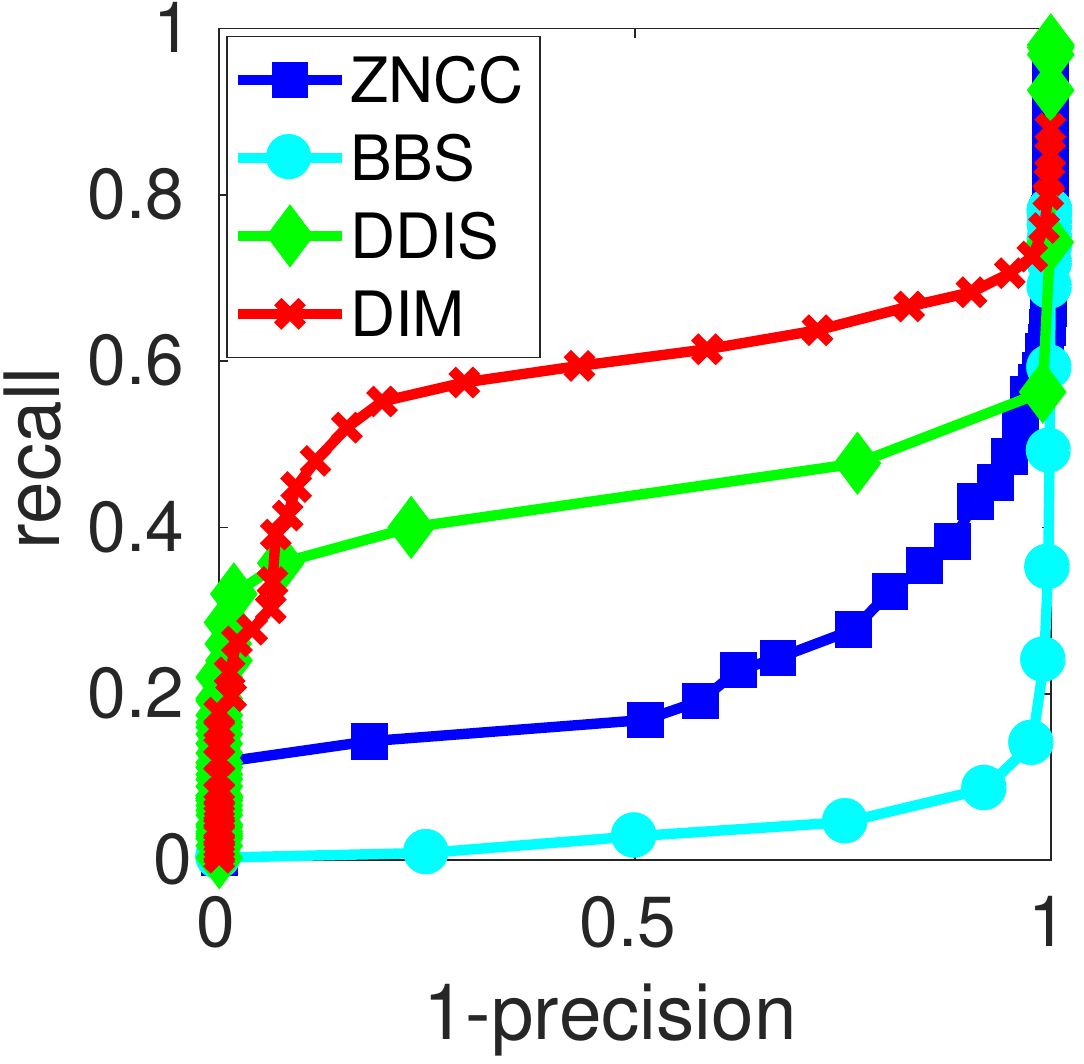}}
  \subfigure[]{\includegraphics[scale=0.4,trim=0 0 0 0, clip]{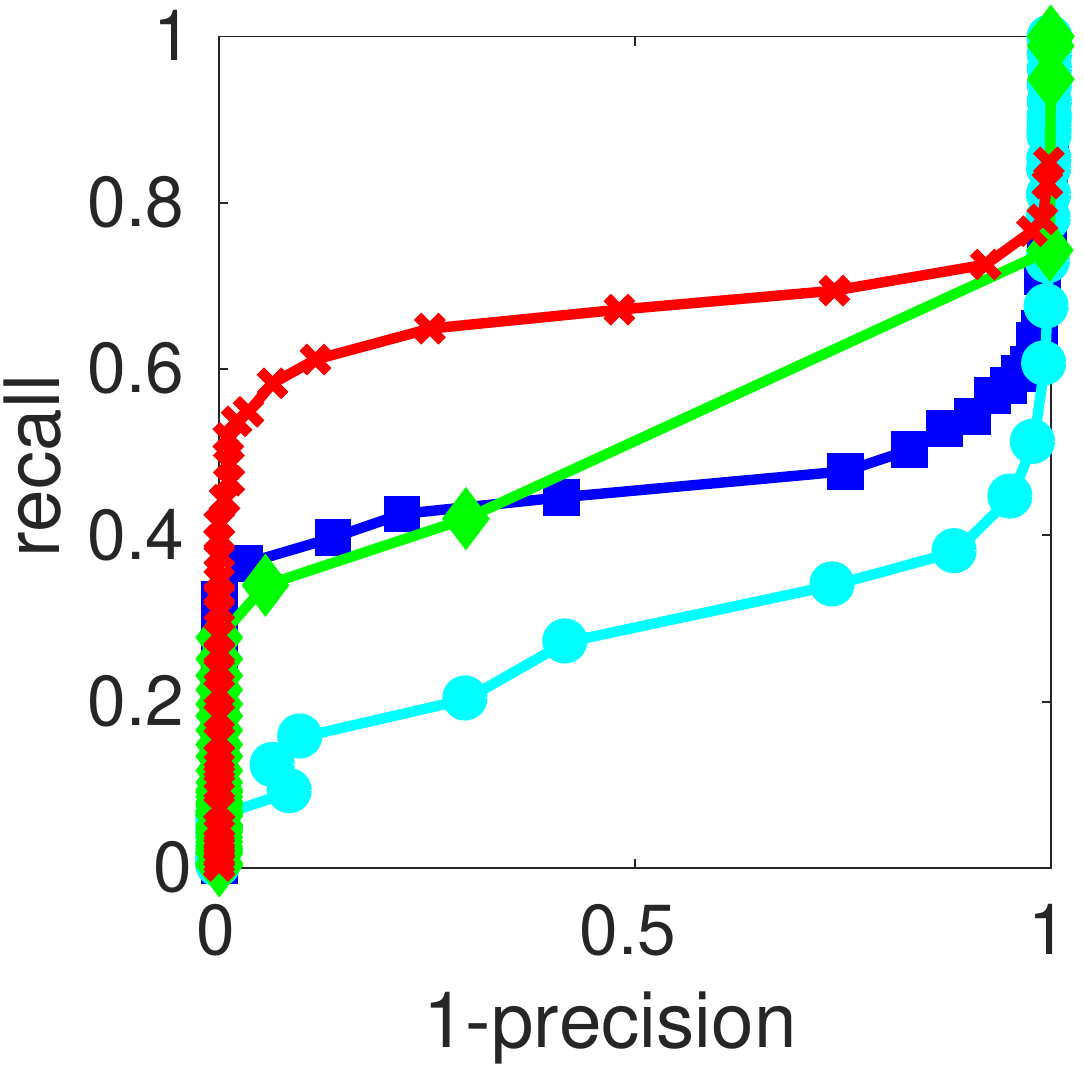}}
  \subfigure[]{\includegraphics[scale=0.4,trim=0 0 0 0, clip]{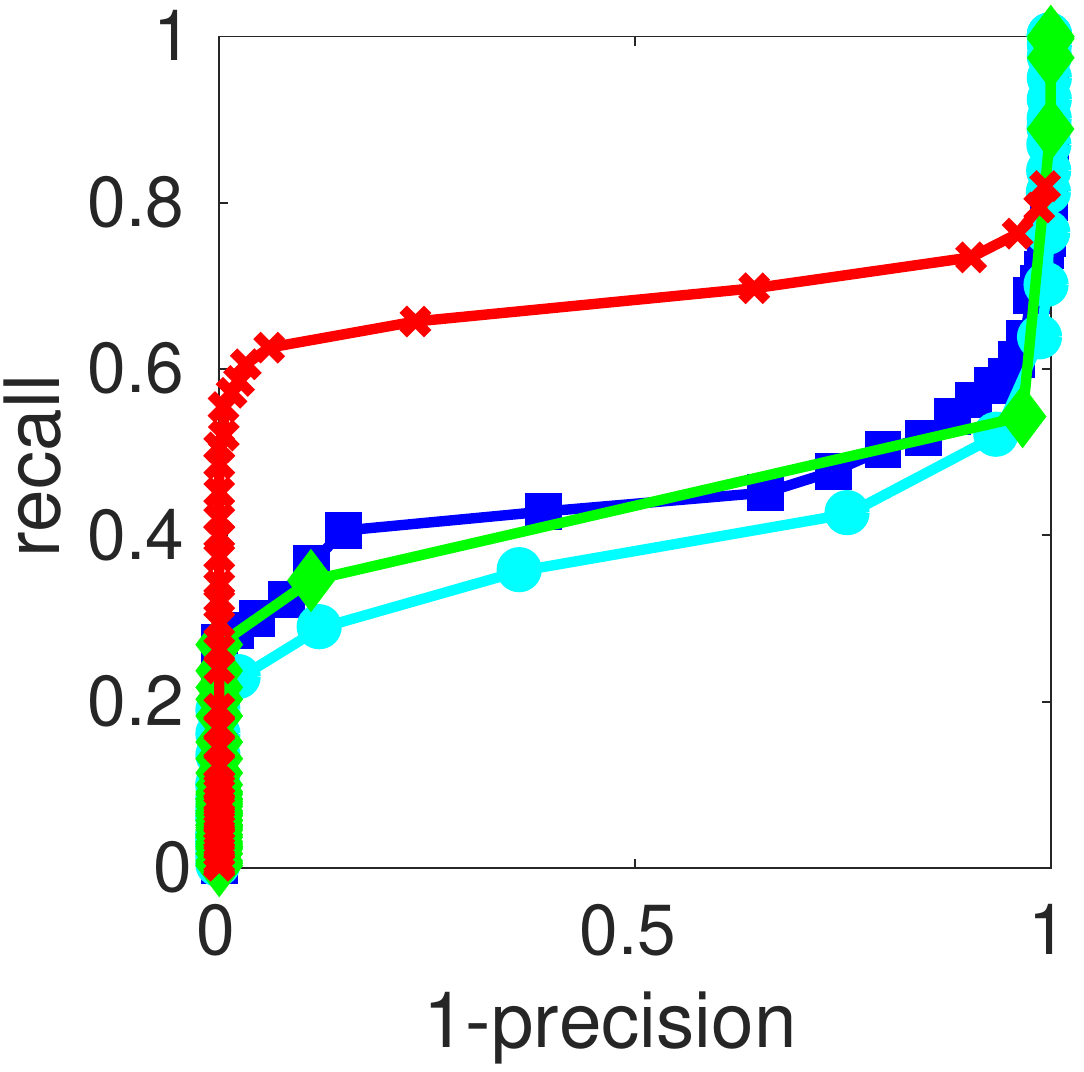}}
\caption{The performance of different algorithms when applied to performing
  template matching in colour images from the Oxford VGG affine covariant
  features dataset (at half size). Each curve shows the trade-off between
  precision and recall for different thresholds applied to the similarity
  values. A match was considered correct if the bounding box overlap between the
  predicted location and the true location was at least 0.5. Results are shown
  for three different sizes of template (a) 17-by-17 pixels, (b) 33-by-33
  pixels, and (c) 49-by-49 pixels.}
\label{fig-vgg_data_precision_recall}
\end{center}
\end{figure*}

The total number of true-positives ($TP$), false-positives ($FP$), and
false-negatives ($FN$) were found for all 70 templates when matched to all 35
query images. These values were then used to calculate recall
($\frac{TP}{TP+FN}$) and precision ($\frac{TP}{TP+FP}$). By varying the global
threshold used to define a match, precision-recall curves were plotted to show
how detection accuracy varied with
threshold. \Autoref{fig-vgg_data_precision_recall} shows the precision-recall
curves for each method for three different sizes of templates. The performance
of each algorithm was summarised by calculating the f-score
($=2\frac{recall.precision}{recall+precision}=\frac{2TP}{2TP+FP+FN}$) at the
threshold that gave the highest value. The f-score measures the best trade-off
between precision and recall. The f-scores for each algorithm are shown in
\autoref{tab-vgg_data_fscores}.

\begin{table}[tbp]
\begin{center}
\dsoff
\begin{tabular}{lllll} \hline
 {\bf Algorithm} & & {\bf f-score} & & \\
   \multicolumn{2}{r}{patch size (pixels):}  & 17-by-17      & 33-by-33      & 49-by-49 \\
\hline
 ZNCC                                 & &      0.2842 &       0.5508  &       0.5493  \\  
 BBS \citep{Dekel_etal15,Oron_etal18} & &      0.0822 &       0.3704  &       0.4579  \\  
 DDIS \citep{Talmi_etal17}            & &      0.5198 &       0.5355  &       0.4959  \\  
 DIM (70 templates)                   & & {\bf 0.6542} & {\bf 0.7230}  & {\bf 0.7513}  \\
\hline
\end{tabular}
\caption{Quantitative comparison of results for different algorithms applied to
  the task of template matching in colour images from the Oxford VGG affine
  covariant features dataset (at half size). Results are given in terms of the
  highest f-score for all 2450 template-image comparisons across the seven
  colour sequences in the dataset (70 templates per image).}
\label{tab-vgg_data_fscores}
\end{center}
\end{table}

From both \autoref{fig-vgg_data_precision_recall} and
\autoref{tab-vgg_data_fscores} it can be seen that the proposed algorithm, DIM,
had the best performance.  The performance of BBS improved as the template size
increased, but at all sizes the performance was well below that of the baseline
method, ZNCC.
This suggests that the similarity values calculated by BBS vary widely in
magnitude such that true and false matches can not be reliably distinguished.
The performance of DDIS was similar for all template sizes, but was only superior
to that of ZNCC at the smallest size.

\subsection{Parameter Sensitivity}
\label{sec-parameter_sensitivity}
The proposed method employs a number of parameters:
\begin{itemize}\itemsep0em 
\item the number of additional templates used;
\item the size of the image patches;
\item the standard deviation, $\sigma$, of the Gaussian mask used to pre-process the images;
\item the value of $\epsilon_1$ in equation \autoref{eq-pcbc_y};
\item the value of $\epsilon_2$ in equation \autoref{eq-pcbc_e};
\item The scale factor, $\lambda$, used to determine the size of the
  elliptical region used to sum neighbouring similarity values;
\item the number of iterations performed by the DIM algorithm.
\end{itemize}
The preceding experiments have already explored the influence of some of these
parameters, specifically, the effects of varying the number of additional
templates is shown in \autoref{fig-bbs_data_num_templates_both}, and the
experiments in \autorefs{sec-correspondence_vgg} and
\ref{sec-template_matching_vgg} have examined the effects of varying the size
of the image patch.  Additional experiments were carried out to measure the
sensitivity of the proposed algorithm to the other parameters. These
experiments applied the algorithm to finding corresponding locations across
105 pairs of colour video frames from the BBS dataset (as in
\autoref{sec-correspondence_bbs}).  In each experiment one parameter was
altered at a time while all other parameters were kept fixed at their default
values. The results are shown in \autoref{tab-parameter_sensitivity}.

It can be seen that the value of $\sigma$ used to pre-process the images could be
increased or decreased by a factor of two, and the algorithm still produced
performance on this task that exceeded the previous state-of-the-art. However,
increasing or decreasing this parameter further had a detrimental effect on
performance. This is not surprising as when $\sigma$ is too large or too small
$\bar{I}$ becomes a poor estimate of the local image intensity: in the limit
$\bar{I}$ becomes equal to the average intensity of the whole image (when
$\sigma$ is very large), or $\bar{I}=I$ (when $\sigma$ is very small). In the
latter case the input to the template matching method becomes an image where all
pixels have a value of zero.

The algorithm was tolerant to large changes in $\epsilon_1$, and
$\epsilon_2$. However, when $\epsilon_2$ was increased by a factor of 10 from
its default value performance deteriorated. This is becomes this large value of
$\epsilon_2$ is significant compared to the values of $\R$ (and $\X$), and this
causes the DIM algorithm to fail to accurately reconstruct its input, and hence,
has non-negligible effects on the steady-state values of $\R$, $\E$ and $\Y$.

Using a very small value of $\lambda$ (which is equivalent to skipping the
post-processing stage described in \autoref{sec-methods_postproc}) resulted in a
AUC of 0.687. As shown in \autoref{tab-parameter_sensitivity} larger values of
$\lambda$ were beneficial, but if $\lambda$ became too large performance
deteriorated. This is because when the summation region is large, small
similarity values scattered across a large region of the image can be summed-up
to produce what appears to be a strong match from multiple, unrelated, weak
matches with the template.

At the end of the first iteration of the DIM algorithm the similarity values are
given by: $\Y_j = \frac{\epsilon_1}{\epsilon_2} \odot \sum_{i=1}^{k}
\left(\w_{ji} \star \X_i\right)$, \ie the cross-correlation between the
templates and the pre-processed image. Hence, unsurprisingly, when only one
iteration was performed, performance was very poor and similar to simple
correlation-based methods like NCC (compare \autoref{tab-parameter_sensitivity}
row ``iterations'' and column ``$\div 10$'' with \autoref{tab-bbs_data_AUC} row
``NCC''). Two iterations was also insufficient for the DIM algorithm to find an
accurate and sparse representation of the image. However, with between 5 and 50
iterations the proposed method produced accurate results that were consistently
equal to or better than those of the previous state-of-the-art.  Performance
deteriorated when a very large number of iterations was performed. However, this
can be offset by increasing the value of $\lambda$. For example, using 100
iterations and $\lambda=0.075$ produced an AUC of 0.666. This can be explained
by the similarity values becoming sparser as the number of iterations increases,
allowing a larger summation region to be used without such a risk of integrating
across unrelated similarity values.

\begin{table}[tbp]
  \begin{center}
\dsoff
\begin{tabular}{lccccccc} \hline
  {\bf Parameter} & {\bf Standard} & \multicolumn{6}{c}{{\bf AUC when value changed by:}}\\
                  & {\bf  Value}   & $\div 10$ & $\div 5$ & $\div 2$ & $\times 2$ & $\times 5$ & $\times 10$ \\
\hline
$\sigma$ (see \autoref{sec-methods_preproc}) & 0.5min(w,h) & 0.536 & 0.597 & 0.681 & 0.674 & 0.609 & 0.554\\
$\epsilon_1$ (see \autoref{eq-pcbc_y}) & $\frac{\epsilon_2}{max\left(\sum_j v_{ji}\right)}$ & 0.688 & 0.688 & 0.690 & 0.690 & 0.689 & 0.688 \\
$\epsilon_2$ (see \autoref{eq-pcbc_e}) & 0.01 & 0.690 & 0.690 & 0.690 & 0.686 & 0.691 & 0.624\\
$\lambda$ (see \autoref{sec-methods_postproc}) & 0.025 & 0.695 & 0.695 & 0.695 & 0.682 & 0.682 & 0.636\\
iterations (see \autoref{sec-methods_matching})   & 10 & 0.451 & 0.593 & 0.687 & 0.668 & 0.644 & 0.632 \\
\hline
\end{tabular}
\caption{Evaluation of the sensitivity of the proposed algorithm to its
  parameter values. Note, w and h stand for the width and height of the
  template. Performance was evaluated using the AUC produced for the task of
  finding corresponding locations in 105 pairs of colour video frames from the BBS
  dataset (\ie using the same procedure as used to generate the result shown in
  \autoref{tab-bbs_data_AUC}). Using the standard parameter values the AUC is
  equal to 0.690.}
\label{tab-parameter_sensitivity}
\end{center}
\end{table}

\subsection{Computational Complexity}
\label{sec-speed}

The focus of this work was to develop a more accurate method of template
matching. Computational complexity was therefore not of prime concern. However,
for completeness, this section provides a comparison of the computational
complexity of DIM.
  
To calculate the cross-correlation or convolution of a $M$-by$N$ pixel image
with a $m$-by-$n$ pixel template, requires $mn$ multiplications at each image
pixel, so the approximate computational complexity is $O(MNmn)$. In the DIM
algorithm, to avoid edge effects, the image is padded by $2m$ in width and $2n$
in height. In this case, the computational complexity of 2D cross-correlation is
$O((M+2m)(N+2n)mn)$.

In the DIM algorithm, 2D convolution is applied to calculate the values of $\R$
for each channel (see \autoref{eq-pcbc_r}) and 2D cross-correlation is used to
calculate the values of $\Y$ for each template (see \autoref{eq-pcbc_y}). Both
these updates are performed at each of $i$ iterations. So if there are $c$
channels and $t$ templates, then the complexity is approximately
$O(2(M+2m)(N+2n)mncti)$.  Added to this is the time taken to compute the
element-wise division (see \autoref{eq-pcbc_e}) and multiplication (see
\autoref{eq-pcbc_y}) operations at each iteration, which has a complexity of
$O(MN(c+t)i)$. However, this is negligible compared to the time taken to perform
the cross-correlations and convolutions.

It is well-known that 2D convolution and 2D cross-correlation can be performed
in Fourier space with complexity $O(MNlog(MN))$, assuming $m \le M$ and $n \le
N$. This method is therefore faster when $mn$ is larger than
$log((M+2m)(N+2n))$. Using the Fourier method of calculating the
cross-correlations and convolutions, the complexity of DIM would be
approximately $O(2(M+2m)(N+2n)log(MN)cti)$.  Cross-correlation and
convolution are inherently parallel processes as each output value is
independent of the other output values. Hence, with appropriate multi-core
hardware the computational complexity of 2D cross-correlation and 2D convolution
becomes $O(mn)$. With such parallel computation, the computational complexity of
DIM would be approximately $O(2mni)$, as the value of $\R$ across all channels
and the values of $\Y$ for all templates could also be calculated in parallel.

This compares to the complexity of ZNCC which is $O(MNmnct)$ (or $O(mn)$ on
parallel hardware); BBS which is $O(MNm^2n^2ct)$ \citep{Oron_etal18}; and DDIS
which is $O(9mnlog(mn)t+MNmnt+MNt(c+log(mn)))$ \citep{Talmi_etal17}.  To compare
the real-world performance, execution times for each algorithm were recorded on
a computer with Intel Core i7-7700K CPU running at 4.20GHz. This machine ran
Ubuntu GNU/Linux 16.04 and MATLAB R2017a. All code was written in MATLAB. For
DDIS faster, compiled, code is available to reduce execution times on machines
running Microsoft Windows. The code for DIM was not compiled for a fair
comparison.  The total time taken to perform the experiment described in
\autoref{sec-correspondence_bbs} (\ie to perform template matching across 105
colour image pairs), the time taken to perform the experiment described in
\autoref{sec-correspondence_vgg} with 17-by-17 pixel templates (\ie to match 25
templates across 40 image pairs), and the time taken to perform the experiment
described in \autoref{sec-template_matching_vgg} with 17-by-17 pixel templates
(\ie to match 70 templates to 35 images) are shown in
\autoref{tab-execution_times}. While DIM is not as fast as the simple, baseline,
method it is the fastest of the other methods while also being the most
accurate. It also has the potential to be much faster if implemented on
appropriate parallel hardware.

\begin{table}[tbp]
\begin{center}
\dsoff
\begin{tabular}{lrrr} \hline
{\bf Algorithm} & \multicolumn{3}{l}{{\bf Execution Time (s)}} \\
\hline
ZNCC                                     & {\bf 15 (0.14)} & {\bf 33 (0.03)} & {\bf 59 (0.02)}\\
BBS \citep{Dekel_etal15,Oron_etal18}     & 2193    (20.89) & 724     (0.72) & 1512     (0.62)\\ 
DDIS \citep{Talmi_etal17}                & 4802    (45.73) & 17391  (17.39) & 40102   (16.37)\\
DIM                                      & 938      (8.93) & 209     (0.21) & 1366     (0.56)\\ 
\hline
\end{tabular}
\caption{Comparison of the execution times of different algorithms. The first
  column of times show the total time taken when the algorithms where applied to
  the task of finding corresponding locations in 105 pairs of colour video frames
  (\ie to obtain the results shown in \autoref{fig-bbs_data_success}). The
  second column of times are for the task of finding 25 corresponding points in
  each pair of images from the Oxford VGG affine covariant features dataset, at
  one-half size, using 17-by-17 pixel templates (\ie to obtain the results shown
  in the first column of \autoref{fig-vgg_data_success}). The third column of
  times are for the algorithms when applied to the task of matching 70 17-by-17
  pixel templates to 35 colour images from the Oxford VGG affine covariant
  features dataset, at half size (\ie to obtain the results shown in
  \autoref{fig-vgg_data_precision_recall}a). The values in brackets are the
  average times taken to compare one template with one image in each task.}
\label{tab-execution_times}
\end{center}
\end{table}

\section{Conclusions}
\label{sec-discussion}

This article has evaluated a method of performing template matching that is
shown to be both accurate and tolerant to differences in appearance due to
viewpoint, variations in background, non-rigid deformations, illumination,
blur/de-focus and JPEG compression. This advantageous behaviour is achieved by
causing the templates to compete to match the image, using the existing DIM
algorithm \citep{Spratling_etal09,Spratling17a}. Specifically, the competition
is implemented as a form of probabilistic inference known as explaining away
\citep{Kersten_etal04,LochmannDeneve11,Spratling_dim-learning,Spratling_etal09}
which causes each image element to only provide support for the template that is
the most likely match. Explaining away produces a sparse array of similarity
values in which the peaks are easily identified, and in which similarity values
that are reduced in magnitude by differences in appearance are still distinct
from those similarity values at non-matching locations. Using a variety of
tasks, the proposed method was shown to out-perform traditional template
matching, and recent state-of-the-art methods
\citep{Talmi_etal17,Dekel_etal15,Oron_etal18,Kat_etal18,Kim_etal17,ZagoruykoKomodakis15,ZagoruykoKomodakis17}.

Specifically, the proposed method was compared to the BBS algorithm
\citep{Dekel_etal15,Oron_etal18}, and several other recent methods
\citep{ZagoruykoKomodakis15,ZagoruykoKomodakis17,Kat_etal18,Kim_etal17,Talmi_etal17},
using the same dataset and experimental procedures defined by the authors of the
BBS algorithm. This task required target objects from one frame of a colour
video to be located in a subsequent frame. Changes in the appearance of the
target were due to variations in camera viewpoint or the pose of the target,
partial occlusions, non-rigid deformations, and changes in the surrounding
context, background, and illumination. On this dataset the proposed algorithm
produced significantly more accurate results than the BBS algorithm, and more
recent algorithms that have been applied to the same dataset
\citep{ZagoruykoKomodakis15,ZagoruykoKomodakis17,Kat_etal18,Kim_etal17,Talmi_etal17}.
Furthermore, using the Oxford VGG Affine Covariant Features Dataset
\citep{MikolajczykSchmid05,Mikolajczyk_etal05} it was shown that these findings
generalise to other tasks and other images. In this second set of images,
changes in the appearance of the target were due to variations in camera
viewpoint, illumination/exposure, blur/de-focus, and JPEG compression.  The
proposed method considerably outperformed some recently proposed
state-of-the-art methods \citep{Dekel_etal15,Oron_etal18,Talmi_etal17} on these
additional experiments.
  
The present results demonstrate that the proposed method is tolerant to a range
of factors that cause differences between the template and the target as it
appears in the query image. However, it is only weakly tolerant to changes in
appearance caused by viewpoint (\ie changes in perspective, orientation, and
scale).  The tolerance of DIM to viewpoint changes could, potentially, be improved
using a number of techniques.

Firstly, by using additional templates representing transformed versions of the
searched-for image patch. For example, to recognise a patch of image at a range
of orientations it would be possible to include additional templates showing
that patch at different orientations. The final similarity measure at each image
location would then be determined by finding the maximum of the similarity
values calculated for each individual template representing transformed versions
of the same template at that location. Additional, unreported, experiments have
shown that this method works well. However, to deal with an unknown
transformation between the template and the query image it is necessary to use a
large number of affine transformed templates showing many possible combinations
of changes in scale, rotation and shear, and hence, this method is
computationally expensive and not very practical.

Secondly, it would be possible to split templates into multiple sub-templates.
The sub-templates could be matched to the image, using DIM, and each one could
vote for the location of the target.  By allowing some tolerance in the range of
sub-template locations that vote for the same target location, this method could
provide additional tolerance to changes in appearance, particularly changes
caused by image shear and changes in perspective. Essentially, this method
would perform template matching using an algorithm analogous to the implicit
shape model \citep[ISM; ][]{Leibe_etal08}, which employs the generalised Hough
transform \citep{Ballard81,DudaHart72}. Both the sub-template matching and the
votong processes could be implemented using the DIM algorithm
\citep{Spratling17a,Spratling16c}.

Thirdly, it would be possible to apply the method to a different feature-space,
one in which the features were tolerant to changes in appearance.  It has been
shown that for other methods significant improvement in performance can be
achieved by applying the method to a feature-space defined by the output of
certain layers in a CNN \citep{Kim_etal17,Kat_etal18,Talmi_etal17}. For example,
\citet{Kim_etal17} showed that applying NCC to features extracted by a CNN, in
comparison to using NCC to compare colour intensity values, produced an increase
of 0.15 in the AUC for the experiment described in
\autoref{sec-correspondence_bbs}.
An obvious direction for future work on the proposed algorithm is to apply it to
a similar feature-space extracted by a deep neural network.

In terms of practical applications, the proposed method
has already been applied, as part of a hierarchical DIM network, to object
localisation and recognition \citep{Spratling17a} and to the low-level task of
edge-detection \citep{Spratling13a,WangSpratling16b}.  Future work might also
usefully explore applications of the proposed method to stereo correspondence,
3D reconstruction, and tracking. To facilitate such future work all the code
used to produce the results reported in this article has been made freely available.

\bibsep=0pt

\end{document}